\pdfoutput=1
\documentclass[11pt,letterpaper]{trackrl_arxiv}

\usepackage{pifont}
\usepackage{url}
\usepackage{wrapfig}
\usepackage{algorithm}
\usepackage{algorithmic}

\usepackage{amsmath,amsfonts,bm}

\def\eqref#1{equation~\ref{#1}}

\def\1{\bm{1}}

\DeclareMathAlphabet{\mathsfit}{\encodingdefault}{\sfdefault}{m}{sl}
\SetMathAlphabet{\mathsfit}{bold}{\encodingdefault}{\sfdefault}{bx}{n}

\definecolor{linkblue}{RGB}{41, 98, 155}
\definecolor{citegreen}{RGB}{85, 100, 40}
\hypersetup{colorlinks=true, linkcolor=linkblue, urlcolor=linkblue,
            citecolor=citegreen, filecolor=linkblue}

\newtcolorbox{promptbox}{colback=gray!5, colframe=black!60, boxrule=0.5pt, arc=2pt,
  left=5pt, right=5pt, top=4pt, bottom=4pt, breakable}
\makeatletter\def\verbatim@font{\small\ttfamily}\makeatother

\title{Trackable Skill Library Lifecycle Management\\ for Agentic Reinforcement Learning}

\author[1,*]{Xi Zhu}
\author[1,*]{Yuanlin Duan}
\author[2]{Changyi Yang}
\author[1]{Kai Mei}
\author[1]{Riyang Bao}
\author[3]{Shichao Zhu}
\author[1]{Dimitris N. Metaxas}

\renewcommand\authorlist{%
  \authorformat[1,*]{Xi Zhu}\quad
  \authorformat[1,*]{Yuanlin Duan}\quad
  \authorformat[2]{Changyi Yang}\quad
  \authorformat[1]{Kai Mei}\\[2mm]
  \authorformat[1]{Riyang Bao}\quad
  \authorformat[3]{Shichao Zhu}\quad
  \authorformat[1]{Dimitris N. Metaxas}}

\affiliation[1]{Rutgers University}
\affiliation[2]{Carnegie Mellon University}
\affiliation[3]{TikTok Inc.}

\abstract{%
Skill-augmented agentic reinforcement learning (RL) has emerged as an effective approach to improving LLM agents on long-horizon tasks. Despite this progress, individual skills may follow distinct trajectories as the policy evolves, with some becoming redundant after the model internalizes their capabilities, others remaining valuable as external scaffolding, and still others turning harmful. However, existing skill library systems fail to identify each skill's true contribution due to free-riding among co-retrieved skills, difficulty confounding across tasks, and the absence of skill-specific counterfactual evidence, which prevents reliable skill lifecycle management. We propose TrackRL, an agentic RL framework in which skill tracking drives full lifecycle management of the skill library. Specifically, within each GRPO group, we turn grouped sampling into an online randomized controlled experiment. Independent random injection and within-group contrasts on the same task isolate each skill's true contribution from co-retrieved skills and task difficulty, and reliability-weighted aggregation accumulates this evidence across groups. As evidence accumulates, we continuously track skill utility, which measures the skill's causal contribution as external scaffolding, and model mastery, which reflects how far the corresponding capability has been internalized. Along this line, we formulate skill lifecycle management as an evidence-gated state-transition process that governs each skill from generation and admission through protected incubation to retention, retirement, or removal, while a soft capacity bound controls library growth. Extensive experiments show that TrackRL outperforms recent skill-augmented RL baselines in both library-assisted and skill-free evaluation, with further analyses revealing how skill value shifts during policy learning and how evidence-driven state transitions shape skill library evolution. Our code is now available at \url{https://github.com/xizhu1022/TrackRL}.
}

\begin{document}

\maketitle
{\renewcommand{\thefootnote}{}\footnotetext{$^{*}$Equal contribution.\quad Contact: \email{xi.zhu@rutgers.edu}}}

\section{Introduction}
Agentic reinforcement learning (RL) has substantially improved LLM agents on long-horizon tasks, including household planning \citep{shridhar2021alfworld}, web navigation \citep{yao2022webshop}, and interactive coding tasks \citep{trivedi2024appworld}. Existing skill-augmented approaches optimize model policies through RL while distilling experience into reusable skills stored in an external library \citep{xia2026skillrl, yao2026skillrise}. As the skill library grows, relevant skills are retrieved as context to guide subsequent policy updates, enabling policy--library co-evolution \citep{shi2026skill1, li2026skillgraph}.

This co-evolution dynamically couples the model's internal policy with its external skill library. A skill derived from reflection on a failed trajectory fills a capability gap at one stage of training, but that gap may narrow or shift as the policy evolves. As RL continually reshapes model capabilities and, in turn, revalues existing skills \citep{yu2026forget, yu2026skill}, an effective skill library should evolve with the policy it supports, rather than be constructed and maintained in isolation \citep{he2026reskill}.

\begin{wrapfigure}{r}{0.5\linewidth}
\vspace{-4mm}
\centering
\includegraphics[width=\linewidth]{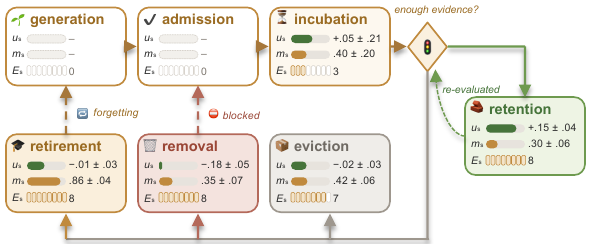}
\caption{\textbf{Skill lifecycle.} Tracked skill utility, model mastery, and evidence drive the lifecycle of each skill. Retained skills serve as external scaffolding, whereas retired skills have been internalized by the policy.}
\label{fig:motivation}
\end{wrapfigure}

Figure~\ref{fig:motivation} illustrates the distinct evolutionary trajectories of individual skills. As the model policy evolves, some skills lose marginal value and become redundant once the model internalizes the capability \citep{xiao2026uimem, xia2026unimem}. Others remain valuable as external scaffolding for long-tail cases or hard-to-internalize knowledge \citep{wang2024workflow, suzgun2025dynamic}, while some become harmful due to reflection errors or policy changes \citep{liu2026interference, zhang2026consistencygate}. To manage these divergent trajectories, a continually learning agent must track each skill's current utility and the model's mastery of the corresponding capability, and use the accumulated evidence to govern each skill's full lifecycle, keeping the skill library aligned with the evolving model policy \citep{pu2026skillops, lin2026muse}.

However, today's skill library systems cannot identify what an individual skill contributes or whether the current model still needs it. First, a successful rollout typically relies on multiple retrieved skills, yet its episode-level outcome is attributed indiscriminately to all of them \citep{tu2026d2skill, li2026attrimem}, allowing non-contributing skills to free-ride on those that actually drive success. This ambiguity can be amplified when the skill library contains both task-level planning skills and step-level execution skills, as each action draws on a different set of step-level skills, but all such sets are still evaluated against a single episode-level outcome \citep{wang2026bipace}. Second, because skills are retrieved rather than randomly assigned \citep{xia2026skillrl, wang2023voyager}, a skill's observed utility depends partly on the difficulty of the tasks it serves, a phenomenon we term difficulty confounding. Successes on easy tasks may inflate its apparent contribution, while failures on hard tasks may obscure meaningful assistance. Most critically, existing pipelines lack the skill-specific counterfactual evidence needed to estimate a skill's marginal utility and the model's mastery of the corresponding capability \citep{hu2026persistent}, leaving its evolving value difficult to assess. We therefore need a debiased, longitudinal estimator that tracks each skill's utility across task types and co-retrieved skill sets, together with the model's mastery of the corresponding capability, to provide an evidence-based foundation for skill library lifecycle management.

Even as skill-level evidence accumulates, existing frameworks lack a principled mechanism for translating it into reliable skill lifecycle decisions. On the one hand, they address capacity constraints through utility-based pruning, without protecting skills whose limited evidence may conceal latent value during their incubation periods. On the other hand, they fail to distinguish the diversity of skill lifecycle states and the complexity of transitions between them. Each skill follows a distinct life course: born from reflection, incubated as evidence accumulates, evolves through repeated retrieval and use, and may eventually retire once the model internalizes the corresponding capability \citep{deng2024explicit}, persist as an external scaffold \citep{suzgun2025dynamic}, or be removed if it becomes harmful. 
We therefore need trackable, state-driven skill lifecycle management that leverages skill utility and model mastery to identify each skill's evolving state, gates every transition on sufficient evidence, and guides each skill from generation and admission through protected incubation to eventual retention, retirement, or harmful removal.

We propose \textbf{TrackRL}, an agentic RL framework in which skill tracking drives full lifecycle management of the skill library, closing the loop of policy--library co-evolution. Building on the standard skill-augmented RL pipeline, TrackRL turns the grouped sampling of agentic RL into an online randomized controlled experiment by independently randomizing the injection of each selected skill within a group. The resulting within-group contrasts yield two quantities that jointly characterize the evolving state of a skill, namely skill utility, its causal contribution as external scaffolding, and model mastery, the extent to which its capability has been internalized, while addressing free-riding and difficulty confounding by design. Grounded in this evidence, we formulate skill library lifecycle management as an evidence-gated state-transition process. Using tracked utility, mastery, and accumulated evidence, we govern each skill from generation and admission through protected incubation to retention, retirement, or removal under a soft capacity, with each transition supported by sufficient evidence. 

Experiments on ALFWorld and AppWorld show that TrackRL outperforms recent skill-augmented RL baselines in both skill-assisted and skill-free evaluation, and further analyses reveal how skill value shifts as the policy learns and how evidence-driven transitions shape the evolution of the library.
Extensive experiments show that TrackRL outperforms recent skill-augmented RL baselines in both library-assisted and skill-free evaluation. Further analyses demonstrate how skill value shifts as the policy learns and how evidence-driven state transitions shape library evolution.

\section{TrackRL}
\label{sec:method}

\begin{figure}[!t]
\centering
\vspace{-6mm}
\includegraphics[width=0.98\linewidth]{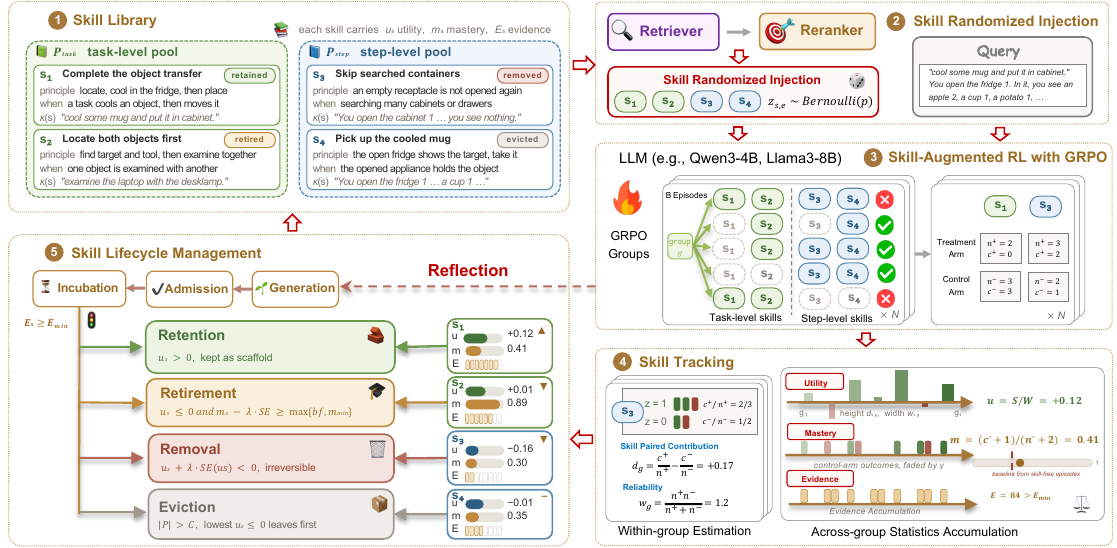}
\caption{\textbf{Overview of TrackRL.} We maintain a skill library of task-level and step-level skills, where we randomize their injection within GRPO groups to create per-skill treatment and control arms. From these comparisons, we track skill utility, model mastery, and evidence to guide skill lifecycle management. The updated library then supports subsequent policy optimization, closing the loop of policy--library co-evolution.}
\vspace{-2mm}

\label{fig:overview}
\end{figure}

TrackRL couples policy optimization with evidence-guided skill library management, as illustrated in Figure~\ref{fig:overview}. Section~\ref{sec:tracking} introduces skill tracking, which uses randomized skill injection and within-group contrasts to estimate skill utility under the current policy and the model's mastery of the corresponding capability. Section~\ref{sec:lifecycle} describes how the accumulated evidence guides skill generation, admission, retention, retirement, and removal, while protecting skills in incubation and controlling library growth. The updated library supports subsequent policy training, and skills are continually reassessed as the policy evolves, closing the loop between policy learning and library management.

\subsection{Skill-Augmented RL with Skill Tracking}
\label{sec:tracking}

\textbf{Skill Organization.} LLM agents require both task-level planning and step-level execution to solve long-horizon tasks. Following prior work \citep{tu2026d2skill, agashe2024agents, wang2025mobileagente}, we adopt a hierarchical skill organization and maintain two initially empty textual skill pools, a task-level skill pool $\mathcal{P}_{\mathrm{task}}$ and a step-level skill pool $\mathcal{P}_{\mathrm{step}}$. Specifically, task-level skills generalize across tasks with similar planning structures and guide episode-level decisions, whereas step-level skills generalize across similar local situations and guide action selection from the current observation.

\textbf{Skill Retrieval and Reranking.} From the skill library, the agent must select a few skills that are both relevant to the current query and useful under the current model policy. Accordingly, we retrieve candidates by semantic similarity and rerank them with accumulated usage evidence. Each skill $s$ is associated with a retrieval key $\kappa(s)$, a short description of the context in which it applies. We retrieve task-level skills once at the beginning of each episode using the task description as query $x$, and step-level skills at every turn the task description concatenated with the current observation. Formally, we encode the query and retrieval keys through a text embedding model, and retrieve at most $m$ candidates from the corresponding pool $\mathcal{P}$,
\begin{equation}
\mathcal{R}(x;\mathcal{P}) \;=\; \operatorname{Top}_{m}\!\big\{\, s \in \mathcal{P} \;\big|\; \operatorname{sim}\big(x, \kappa(s)\big) \geq \delta \,\big\} ,
\label{eq:retrieve}
\end{equation}
where $\operatorname{sim}(x,\kappa(s))$ is the cosine similarity and $\delta$ is a threshold. Retrieval identifies semantically relevant skills, but relevance alone does not show whether a skill helps or harms the current policy. Thus, we rerank the candidates by combining semantic relevance and usage evidence,
\begin{equation}
\operatorname{score}(s) \;=\; \alpha\, \widetilde{\operatorname{sim}}\big(x, \kappa(s)\big) \;+\; (1-\alpha) \left[\, u_s + \eta \sqrt{ \frac{\log\big(1+N_{\mathcal{P}}\big)}{1+N_s} } \,\right] ,
\label{eq:score}
\end{equation}
where $\widetilde{\operatorname{sim}}=(\operatorname{sim}+1)/2$ maps cosine similarity to $0,1$, $u_s$ is the current utility estimate, $N_s$ counts episodes in which $s$ was selected, and $N_{\mathcal{P}}$ is the total selection count across the pool. The coefficient $\alpha$ balances semantic relevance against usage evidence, while $\eta$ controls an exploration bonus that decreases with exposure, giving less frequently selected skills opportunities to be evaluated before their utility estimates become reliable \citep{auer2002finite}.
Finally, we select the $K$ highest-scoring candidates for prompt injection.
We keep task-level skills active throughout the episode, whereas we append step-level skills after the current observation to guide only the next action generation.

\textbf{Skill-Augmented Policy Optimization.} We optimize the policy using standard GRPO \citep{shao2024deepseekmath}. The policy $\pi_\theta$ solves each task through multi-turn interaction, with assigned skills provided as context at each turn. Each episode $e$ produces a binary outcome $y_e \in {0,1}$ and receives reward $R_e = y_e - \nu\phi_e$, where $\phi_e$ is the fraction of invalid actions. Unlike methods that give additional rewards for skill-conditioned episodes \citep{xia2026skillrl, tu2026d2skill}, we omit such rewards, as they may encourage reliance on skills rather than task completion. At each training step, we sample $M$ tasks, generate a group $g$ of $B$ episodes per task using the behavior policy $\pi_{\mathrm{old}}$, and maximize
\begin{equation}
\mathcal{J}(\theta) = \mathbb{E}_{g}\Big[ \tfrac{1}{B} {\textstyle\sum_{e \in g}} \tfrac{1}{L_e} {\textstyle\sum_{k=1}^{L_e}} \min\big( \rho_{e,k} A_e,\ \operatorname{clip}( \rho_{e,k}, 1{-}\epsilon, 1{+}\epsilon )\, A_e \big) \Big] - \beta\, D_{\mathrm{KL}}\big( \pi_\theta \,\|\, \pi_{\mathrm{ref}} \big),
\label{eq:grpo}
\end{equation}
with
\begin{equation*}
\rho_{e,k} = \frac{\pi_\theta(o_{e,k} \mid h_{e,k})}{\pi_{\mathrm{old}}(o_{e,k} \mid h_{e,k})},
\qquad
A_e = \frac{R_e - \operatorname{mean}_{e' \in g} R_{e'}}{\operatorname{std}_{e' \in g} R_{e'}},
\end{equation*}
where $L_e$ is the number of generated tokens in episode $e$, $h_{e,k}$ is the context preceding token $o_{e,k}$, and $\epsilon$ and $\beta$ set the clipping range and the KL weight. In standard GRPO, rollouts within a group share the same task and policy, forming repeated trials under otherwise matched conditions. In skill-augmented RL, this structure motivates us to vary skill assignments across these rollouts so that their episode-level outcomes can reveal each skill's contribution.

\textbf{Skill Randomized Injection.} To identify individual skill contributions from episode-level outcomes, we independently randomize the injection of each selected skill across episodes within a GRPO group. Accordingly, for each skill $s$ selected for episode $e$, we sample
\begin{equation}
z_{s,e} \sim \operatorname{Bernoulli}(p),
\qquad
z_{s,e} \perp z_{s',e}
\quad \forall s \neq s'.
\label{eq:randomized_injection}
\end{equation}
For each skill $s$ selected after retrieval and reranking, episodes with $z_{s,e}=1$ receive the skill and form the treatment arm, whereas those with $z_{s,e}=0$ withhold it and form the control arm. Consequently, each group becomes a set of concurrent randomized controlled experiments over multiple selected skills without additional rollouts. 
Randomized injection thus compares the two arms under the same task and policy, removing difficulty confounding, while independent assignments across skills prevent free-riding.
In addition, each group reserves one task-level and one step-level skill-free episode, each omitting all corresponding skills to measure the policy's unaided performance. 
To assign each episode unambiguously to the treatment or control arm for every selected skill, we sample each task-level assignment once at episode onset and each step-level assignment upon first retrieval, reusing the latter whenever the same skill is retrieved again.

\textbf{Skill Utility for the Current Policy.} To quantify each skill's contribution to the current policy, we define its utility $u_s$ as the change in success rate caused by injecting rather than withholding it. We estimate the skill utility $u_s$ from within-group contrasts accumulated over training. For skill $s$ in group $g$, let $n^{+}_{s,g}$ and $n^{-}_{s,g}$ denote the numbers of episodes in the treatment and control arms, and $c^{+}_{s,g}$ and $c^{-}_{s,g}$ their respective numbers of successes. Whenever both arms are nonempty, the group contributes a comparison $d_{s,g}$ with weight $w_{s,g}$,
\begin{equation}
d_{s,g} \;=\; \frac{c^{+}_{s,g}}{n^{+}_{s,g}} - \frac{c^{-}_{s,g}}{n^{-}_{s,g}} , \qquad
w_{s,g} \;=\; \frac{n^{+}_{s,g}\, n^{-}_{s,g}}{n^{+}_{s,g} + n^{-}_{s,g}} .
\label{eq:diff}
\end{equation}
Because the two arms share the same task and policy, $d_{s,g}$ isolates the effect of skill injection from task difficulty. The weight $w_{s,g}$ reflects the reliability of this contrast, giving more weight to comparisons with more episodes and a more balanced split between the arms. As the policy evolves, earlier comparisons become less representative of a skill's current utility, so we discount past evidence by $\gamma \in (0,1]$ and update four running statistics at each training step,
\begin{equation}
\begin{aligned}
W_s &\leftarrow \gamma W_s + \textstyle\sum_{g \in \mathcal{G}_s} w_{s,g} , &\qquad
S_s &\leftarrow \gamma S_s + \textstyle\sum_{g \in \mathcal{G}_s} w_{s,g}\, d_{s,g} , \\
Q_s &\leftarrow \gamma Q_s + \textstyle\sum_{g \in \mathcal{G}_s} w_{s,g}\, d_{s,g}^{2} , &\qquad
G_s &\leftarrow \gamma G_s + |\mathcal{G}_s| ,
\end{aligned}
\label{eq:suffstat}
\end{equation}
where $\mathcal{G}_s$ denotes the groups at the current training step in which both arms are non-empty. $W_s$ accumulates the weights, $S_s$ the weighted comparisons, $Q_s$ their weighted squares, and $G_s$ the effective number of comparisons. These statistics yield the utility estimate and its standard error,
\begin{equation}
u_s \;=\; \frac{S_s}{W_s} , \qquad
\mathrm{SE}(u_s) \;=\; \sqrt{\frac{Q_s - S_s^{2}/W_s}{W_s\,(G_s - 1)}} , \qquad G_s \ge 2 .
\label{eq:utility}
\end{equation}
The estimate $u_s$ captures the skill's current marginal contribution, while $\mathrm{SE}(u_s)$ reflects variation across its comparisons. Since each group contains only $B$ episodes, individual comparisons can be coarse and noisy, making uncertainty essential to avoid relying on a point estimate alone for skill lifecycle decisions. We additionally maintain an undiscounted evidence count $E_s \leftarrow E_s + |\mathcal{G}_s|$.

\textbf{Model Mastery of Skill Capabilities.} Conversely, to assess whether the model has internalized the capability carried by skill $s$, we estimate mastery $m_s$ from the current policy's success rate when $s$ is selected but withheld. Utility alone cannot reveal why a skill contributes little. For example, near-zero utility may indicate either an ineffective skill or a policy that already succeeds without it, with opposite implications for the skill's lifecycle. Control-arm success distinguishes these cases, so we accumulate the corresponding episode and success counts across groups using the same discount,
\begin{equation}
n^{-}_{s} \;\leftarrow\; \gamma\, n^{-}_{s} + \textstyle\sum_{g} n^{-}_{s,g} , \qquad
c^{-}_{s} \;\leftarrow\; \gamma\, c^{-}_{s} + \textstyle\sum_{g} c^{-}_{s,g} .
\label{eq:armcounts}
\end{equation}
Starting with a uniform prior, we use these counts to obtain a posterior $\mathrm{Beta}( c^{-}_{s} + 1,\; n^{-}_{s} - c^{-}_{s} + 1 )$, whose mean and standard deviation give the mastery estimate and its uncertainty,
\begin{equation}
m_s \;=\; \frac{c^{-}_{s} + 1}{n^{-}_{s} + 2} , \qquad
\mathrm{SE}(m_s) \;=\; \sqrt{\frac{m_s\, (1 - m_s)}{n^{-}_{s} + 3}} ,
\label{eq:mastery}
\end{equation}
where $m_s$ starts at a neutral $0.5$ and approaches the observed success rate as evidence accumulates. To keep mastery a reading of the model's capability, we define it only for task-level skills. While success without a task-level skill suggests that the policy has internalized the capability required for the task, success without a step-level skill may instead occur because the skill was nonessential at that step or that later actions or other skills compensated for its absence.

\subsection{Skill Library Full Lifecycle Management}
\label{sec:lifecycle}

\textbf{Skill Generation, Admission, and Protection.} We generate each skill from failure, admit non-redundant candidates, and protect new skills until they accumulate enough evidence for lifecycle decisions. The reflection model extracts task-level and step-level skills from failed trajectories in the groups with the lowest success rates. Since recurring failure patterns can produce near-duplicate skills that dilute the tracking evidence and undermine reliable utility and mastery estimates, deduplication check not only control library size but also preserve reliable statistics. A candidate $s_{\mathrm{new}}$ is admitted only if it is dissimilar to all active skills and all previously removed harmful skills,
\begin{equation}
\max_{s \in \mathcal{P} \cup \mathcal{H}} \mathrm{sim}(s_{\mathrm{new}}, s) \;<\; \delta_{\mathrm{dup}} ,
\label{eq:skill_admission}
\end{equation}
where $\mathcal{H}$ records skills removed as harmful.
Once admitted, a skill starts with empty statistics and enters an incubation period, remaining protected from all exits while $E_s < E_{\min}$ or $\mathrm{SE}(u_s)$ is undefined, regardless of its point estimates. Here, $E_{\min}$ is the minimum evidence threshold.

\textbf{Skill Removal, Retirement, and Retention.} Once sufficient evidence accumulates, we remove harmful skills, retire internalized skills, and retain the rest. Specifically, a skill is deemed harmful and permanently removed once even the upper confidence bound on its utility falls below zero,
\begin{equation}
E_s \ge E_{\min} \;\wedge\; u_s + \lambda\, \mathrm{SE}(u_s) < 0 ,
\label{eq:skill_removal}
\end{equation}
whereas a task-level skill is retired only when its utility is non-positive and even the lower confidence bound on its mastery exceeds both the absolute threshold $m_{\min}$ and the family baseline $b_{f(s)}$,
\begin{equation}
E_s \ge E_{\min} \;\wedge\; u_s \le 0 \;\wedge\; m_s - \lambda\, \mathrm{SE}(m_s) \ge \max\big\{ b_{f(s)},\, m_{\min} \big\} ,
\label{eq:skill_retirement}
\end{equation}
where $\lambda$ sets the confidence level of both one-sided tests and $b_{f(s)}$ denotes the discounted success rate of the task-level skill-free episodes of the same task type. 
Since task difficulty also inflates mastery, this baseline calibrates it to the task family, imposing a higher threshold on skills from easier families.
If model forgetting makes a retired skill necessary again, subsequent failures may trigger its regeneration and readmission as a new candidate, whereas skills removed as harmful remain permanently excluded from the library. As the library converges, skills that retain positive utility without being internalized persist in the skill library as long-term scaffolding.

\textbf{Soft Skill Library Capacity.} The skill library usually requires an upper bound $C$ on the size of each pool to control retrieval cost and storage. However, a hard bound evicts the lowest-ranked skills upon overflow regardless of their evidence, reintroducing the noise-driven pruning prevented by the protection rule. We therefore make $C$ a soft bound. Whenever $|\mathcal{P}| > C$, eligible skills in $\{ s : E_s \ge E_{\min} \wedge u_s \le 0 \}$ are evicted in ascending order of $u_s$, while the pool may remain temporarily above $C$ if this set is exhausted. The capacity rule thus prioritizes the eviction of skills confidently deemed non-beneficial while protecting those still in incubation.

\section{Experiments}
\label{sec:exp}

We conduct experiments on ALFWorld \citep{shridhar2021alfworld} and AppWorld \citep{trivedi2024appworld}, reporting the Success Rate (SR) on ALFWorld and the Task Goal Completion (TGC) and Scenario Goal Completion (SGC) on AppWorld. 
We compare TrackRL against three groups of methods, namely (1) Proprietary LLMs, including Gemini-2.5-Flash and GPT-5.6-Luna under direct prompting, (2) Skill-free models, including the untrained base model and GRPO, and (3) Skill-augmented RL methods, including SkillRL \citep{xia2026skillrl} and D2Skill \citep{tu2026d2skill}. All trained methods use Qwen3-4B-Instruct-2507 as the policy model, and the skill-augmented methods use GPT-5.6-Luna and Gemini-2.5-Flash as reflection models. Full details of the benchmarks, evaluation protocol, baselines, and implementation are provided in Appendix~\ref{app:setup}.

\begin{table*}[!tp]
\centering
\footnotesize
\setlength{\tabcolsep}{4.5pt}
\caption{\textbf{Results on ALFWorld.} SR (\%) in the skill-assisted and skill-free modes on the same checkpoint, with per-class results on the seen split. \emph{Origin} is the base model before RL, Luna and Flash denote the reflection model. "-" marks a method without a skill library. In each setting the best summary result is shown in \textbf{bold}.}
\label{tab:alfworld}
\begin{tabular}{lcccccc|cccc}
\toprule
\multirow{2}{*}{\textbf{Method}} & \multicolumn{6}{c|}{\textbf{Seen, per task class}} & \textbf{Seen} & \textbf{Seen} & \textbf{Unseen} & \textbf{Unseen} \\
 & Pick & Clean & Cool & Look & Heat & Pick2 & \shortstack{skill-\\assisted} & \shortstack{skill-\\free} & \shortstack{skill-\\assisted} & \shortstack{skill-\\free} \\
\midrule
\multicolumn{11}{c}{\textbf{Closed-source LLMs}} \\
\midrule
Gemini-2.5-Flash & 88.6 & 29.6 & 28.0 & 23.1 & 56.2 & 50.0 & - & 50.0 & - & 49.2 \\
GPT-5.6-Luna & 74.3 & 29.6 & 44.0 & 53.8 & 43.8 & 45.8 & - & 50.0 & - & 50.0 \\
\midrule
\multicolumn{11}{c}{\textbf{RL from the base model} (Qwen3-4B-Instruct-2507)} \\
\midrule
Origin & 40.5 & 11.5 & 8.0 & 21.4 & 26.7 & 0.0 & - & 19.3 & - & 27.6 \\
GRPO & 82.9 & 55.6 & 52.0 & 53.8 & 56.2 & 37.5 & - & 58.6 & - & 53.0 \\
SkillRL (Flash) & 11.4 & 11.1 & 12.0 & 7.7 & 12.5 & 8.3 & 11.7 & 20.3 & 10.2 & 20.3 \\
SkillRL (Luna) & 82.9 & 92.6 & 60.0 & 46.2 & 75.0 & 0.0 & 62.1 & 42.1 & 58.2 & 45.5 \\
D2Skill (Flash) & 87.9 & 63.0 & 37.0 & 72.7 & 66.7 & 73.3 & 66.4 & 61.7 & 69.5 & 58.6 \\
D2Skill (Luna) & 80.7 & 83.3 & 18.5 & 41.2 & 63.6 & 50.0 & 57.0 & 52.3 & 54.7 & 52.3 \\
\textbf{TrackRL} (Flash) & 75.0 & 66.7 & 88.0 & 66.7 & 43.8 & 59.1 & 68.8 & 65.6 & 70.3 & 62.5 \\
\textbf{TrackRL} (Luna) & 89.7 & 79.3 & 50.0 & 76.9 & 100.0 & 50.0 & \textbf{78.1} & \textbf{72.7} & \textbf{71.9} & \textbf{64.1} \\
\addlinespace[2pt]
\midrule
\multicolumn{11}{c}{\textbf{RL from the SFT warm up} (Qwen3-4B-Instruct-2507 + Luna SFT, Luna reflection)} \\
\midrule
Origin + SFT & 48.6 & 19.2 & 16.0 & 42.9 & 53.3 & 26.1 & - & 33.6 & - & 40.3 \\
GRPO & 100.0 & 100.0 & 92.0 & 78.6 & 100.0 & 82.6 & - & 93.6 & - & 91.8 \\
SkillRL & 97.1 & 96.3 & 84.0 & 92.3 & 87.5 & 87.5 & 91.4 & 85.0 & 82.8 & 71.6 \\
D2Skill & 100.0 & 100.0 & 90.0 & 100.0 & 93.3 & 94.4 & 96.9 & 93.8 & 96.1 & 91.4 \\
\textbf{TrackRL} & 100.0 & 100.0 & 96.2 & 88.9 & 100.0 & 100.0 & \textbf{98.4} & \textbf{96.1} & \textbf{97.7} & \textbf{95.3} \\
\addlinespace[2pt]
\bottomrule
\end{tabular}
\end{table*}

\subsection{Main Results}
\label{sec:main}
To evaluate whether TrackRL trains stronger agents, we report results on ALFWorld and AppWorld in Tables~\ref{tab:alfworld} and~\ref{tab:appworld}. Each trained method is evaluated in both skill-assisted and skill-free modes using the same checkpoint. We make the following observations. (1) TrackRL outperforms skill-augmented baselines in skill-assisted evaluation under both reflection models. On ALFWorld under cold start, TrackRL achieves 78.1\% and 71.9\% on the seen and unseen splits with GPT-5.6-Luna, compared with 62.1\% and 58.2\% for SkillRL. With Gemini-2.5-Flash, it achieves 68.8\% and 70.3\%, compared with 66.4\% and 69.5\% for D2Skill. These gains persist when the reflection model is held fixed, highlighting the importance of skill attribution and library management beyond the choice of model used to generate skills. The results support our motivation for using skill-specific evidence to guide skill library management. (2) TrackRL also improves the policy's skill-free performance. Starting from the base model on ALFWorld, TrackRL (Luna) achieves 72.7\% and 64.1\% without skill injection, exceeding D2Skill (Flash) at 61.7\% and 58.6\% and GRPO at 58.6\% and 53.0\%. TrackRL (Flash) also exceeds both skill-augmented baselines, achieving 65.6\% and 62.5\%. TrackRL is the only method whose skill-free performance surpasses GRPO under both reflection models. These results indicate that its gains extend to the policy's unaided capabilities, consistent with the role of mastery in tracking whether skills have been incorporated into the policy. (3) TrackRL maintains its advantage from a stronger policy initialization. Starting from the SFT checkpoint, it leads on both ALFWorld splits in both evaluation modes. In skill-assisted evaluation, TrackRL achieves 98.4\% and 97.7\%, compared with 96.9\% and 96.1\% for D2Skill. In skill-free evaluation, it achieves 96.1\% and 95.3\%, compared with 93.8\% and 91.4\% for D2Skill and 93.6\% and 91.8\% for GRPO. Although the margins narrow with stronger initialization, the advantage remains consistent across both splits and modes. This pattern is consistent with the motivation for reassessing each skill's value against the current policy as its capabilities improve. (4) TrackRL's advantage extends to AppWorld, which involves code-based tasks rather than the household tasks of ALFWorld. From the base model, no baseline outperforms TrackRL (Luna) on TGC or SGC, including 23.2\% TGC-N with skill assistance and 22.6\% without it, compared with 15.5\% for GRPO. From the SFT-initialized policy, it leads on nearly all of them, achieving 33.3\% and 34.5\% TGC-N in the two modes, respectively. The consistent gains in both skill-assisted and skill-free evaluation across the two benchmarks support the effectiveness of TrackRL's skill attribution and library management across different task environments.

\newcommand{\val}[1]{\textcolor{gray}{#1$^{\dagger}$}}
\begin{table*}[!tp]
\centering
\footnotesize
\setlength{\tabcolsep}{6pt}
\caption{\textbf{Results on AppWorld.} TGC and SGC (\%) on the normal (N) and challenge (C) splits in the skill-assisted and skill-free modes on the same checkpoint. Luna and Flash denote the reflection model. In each setting the best summary result is shown in \textbf{bold}.}
\label{tab:appworld}
\begin{tabular}{lcccc|cccc}
\toprule
\multirow{2}{*}{\textbf{Method}} & \multicolumn{4}{c|}{\textbf{Skill-assisted}} & \multicolumn{4}{c}{\textbf{Skill-free}} \\
 & TGC-N & SGC-N & TGC-C & SGC-C & TGC-N & SGC-N & TGC-C & SGC-C \\
\midrule
\multicolumn{9}{c}{\textbf{Closed-source LLMs}} \\
\midrule
Gemini-2.5-Flash & - & - & - & - & 41.1 & 26.8 & 28.3 & 15.1 \\
GPT-5.6-Luna & - & - & - & - & 83.9 & 76.8 & 76.7 & 58.3 \\
\midrule
\multicolumn{9}{c}{\textbf{RL from the base model} (Qwen3-4B-Instruct-2507)} \\
\midrule
Origin & - & - & - & - & 5.9 & 0.0 & 2.4 & 0.7 \\
GRPO & - & - & - & - & 15.5 & 7.1 & 3.8 & 0.7 \\
SkillRL (Flash) & 20.8 & \textbf{10.7} & 6.5 & 1.4 & 16.7 & 8.9 & 5.5 & 1.4 \\
SkillRL (Luna) & 20.8 & 7.1 & 7.0 & 2.2 & 20.8 & 5.4 & 7.4 & 2.2 \\
D2Skill (Flash) & 15.5 & 7.1 & 2.9 & 0.0 & 14.9 & 5.4 & 4.6 & 0.7 \\
D2Skill (Luna) & 17.3 & 7.1 & 2.6 & 0.0 & 20.8 & 10.7 & 4.1 & 0.0 \\
\textbf{TrackRL} (Flash) & 20.2 & 5.4 & 5.5 & 1.4 & 17.9 & 8.9 & 6.5 & 0.7 \\
\textbf{TrackRL} (Luna) & \textbf{23.2} & \textbf{10.7} & \textbf{8.9} & \textbf{2.9} & \textbf{22.6} & \textbf{12.5} & \textbf{10.8} & \textbf{2.9} \\
\midrule
\multicolumn{9}{c}{\textbf{RL from the SFT} (Qwen3-4B-Instruct-2507 + SFT, Luna reflection)} \\
\midrule
Origin + SFT & - & - & - & - & 31.6 & 16.1 & 17.5 & \textbf{7.9} \\
GRPO & - & - & - & - & 30.4 & 16.1 & 13.2 & 5.8 \\
SkillRL & 30.9 & 12.5 & 18.0 & 7.9 & 32.7 & 16.1 & 17.5 & \textbf{7.9} \\
D2Skill & 26.8 & 12.5 & 10.3 & 2.2 & 25.6 & 10.7 & 12.7 & 5.0 \\
\textbf{TrackRL} & \textbf{33.3} & \textbf{14.3} & \textbf{21.1} & \textbf{8.6} & \textbf{34.5} & \textbf{17.9} & \textbf{20.1} & \textbf{7.9} \\
\bottomrule
\end{tabular}
\end{table*}

\subsection{Ablation Study}
\label{sec:ablation}
\begin{figure}[!tp]
\centering
\includegraphics[width=0.88\linewidth]{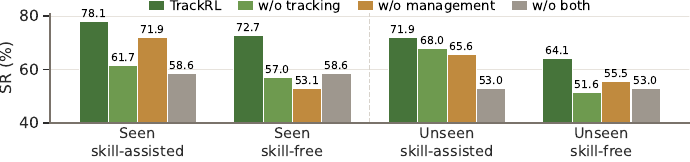}
\caption{\textbf{Ablation on ALFWorld.} Success rate (\%) of TrackRL and its three variants in the four evaluation settings, namely the seen and unseen splits evaluated in the skill-assisted and skill-free modes. The variant w/o both has no library, so its two modes coincide.}
\label{fig:ablation}
\end{figure}

To isolate the contributions of skill tracking and lifecycle management, we compare TrackRL with three variants on ALFWorld. \textbf{(1) w/o tracking} jointly injects or withholds retrieved skills, yielding only bundle-level utility. \textbf{(2) w/o management} admits skills but never retires, removes, or evicts them. \textbf{(3) w/o both} uses no library and reduces to plain GRPO. Figure~\ref{fig:ablation} supports three conclusions. (1) Internalization requires the full mechanism. Only TrackRL improves skill-free SR over GRPO on both splits, reaching 72.7\% versus 58.6\% on seen tasks, while both ablated variants remain near the GRPO level. This suggests that the library strengthens the policy itself when its skills are both tracked and managed. (2) Skill tracking is critical to effective library use. Without per-skill attribution, seen skill-assisted SR falls from 78.1\% to 61.7\%, close to GRPO, as reranking and lifecycle decisions can no longer reliably distinguish useful skills from the rest. (3) Lifecycle management matters most for internalization. Removing all exits lowers seen skill-assisted SR from 78.1\% to 71.9\% but skill-free SR from 72.7\% to 53.1\%, suggesting that indefinitely retained skills may leave the policy reliant on external guidance.

\subsection{Skill Library Growth}
\begin{figure}[!tp]
\centering
\includegraphics[width=0.95\linewidth]{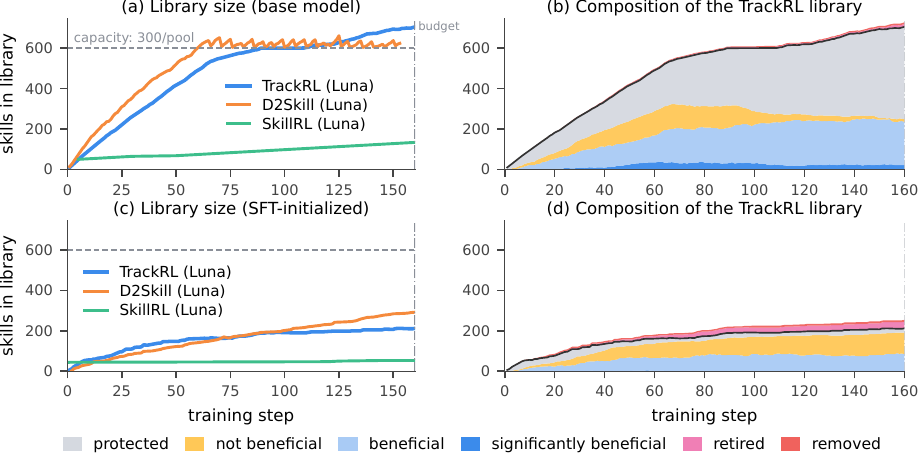}
\caption{\textbf{Skill library growth on ALFWorld.} (a, c) Library size of TrackRL, D2Skill, and SkillRL under cold start and warm start. (b, d) Composition of the TrackRL library by skill state. Significantly beneficial skills have utility exceeding zero by two standard errors.}
\label{fig:library_size}
\end{figure}

To compare library growth under different management schemes, we track the library sizes of SkillRL, D2Skill, and TrackRL and decompose TrackRL by skill state in Figure~\ref{fig:library_size}. Three patterns emerge. (1) After similar initial growth, the skill libraries follow distinct dynamics. SkillRL accumulates skills continuously, D2Skill fluctuates under periodic pruning, and TrackRL initially stabilizes near capacity as evidence-gated exits balance new admissions. (2) TrackRL keeps skills under evaluation and those with positive utility in the library. Cold start from a base model produces more skills in incubation due to frequent early failures, whereas warm start from an SFT-initialized model enables faster evaluation and retirement, keeping the library below capacity. (3) The soft capacity protects skills still in incubation. When no sufficiently evaluated non-beneficial skill remains eligible for eviction, TrackRL allows the library to temporarily exceed capacity. As evidence accumulates, skill library growth is controlled through informed lifecycle decisions.

\subsection{Skills on the Utility--Mastery Plane}
\begin{figure}[!t]
\centering
\includegraphics[width=0.95\linewidth]{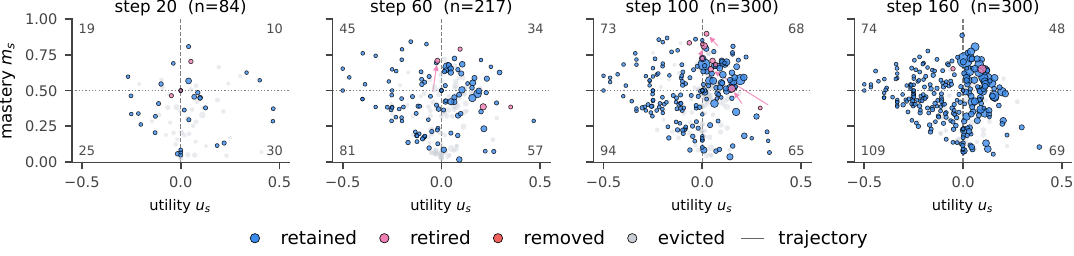}
\caption{\textbf{Task-level skills on the utility--mastery plane.} Each marker represents an active skill positioned by its current utility $u_s$ and mastery $m_s$, with color indicating its eventual lifecycle outcome and size reflecting its accumulated evidence. Corner numbers report the skill count in each quadrant, while arrows trace the earlier positions of skills later retired or removed.} 
\label{fig:phase}
\end{figure}

To examine how utility and mastery jointly inform a skill's fate, we plot every active task-level skill under cold start at four training steps by its current utility and mastery, with color indicating its eventual fate, as shown in Figure~\ref{fig:phase}. We have the following observations. (1) The population moves upward as the policy learns, with the share of skills whose mastery exceeds 0.5 rising from 35\% at step 20 to 41\% at step 160. (2) Utility estimates concentrate near zero as evidence accumulates. Extreme estimates tend to come from skills with few comparisons, whereas skills with more evidence cluster near zero, underscoring the need to consider accumulated evidence rather than a noisy point estimate. (3) Rising mastery is accompanied by declining utility. At the population level, the upper-right quadrant of positive utility and high mastery shrinks from 68 skills at step 100 to 48 at step 160, while the upper-left quadrant remains stable. At the individual level, arrows show skills later retired gaining mastery and approaching zero utility before they leave. Retirement thus appears at high mastery, retention at positive utility, and capacity eviction near zero utility, with evidence determining when a skill can leave the library.

\begin{figure}[!t]
\centering
\includegraphics[width=0.85\linewidth]{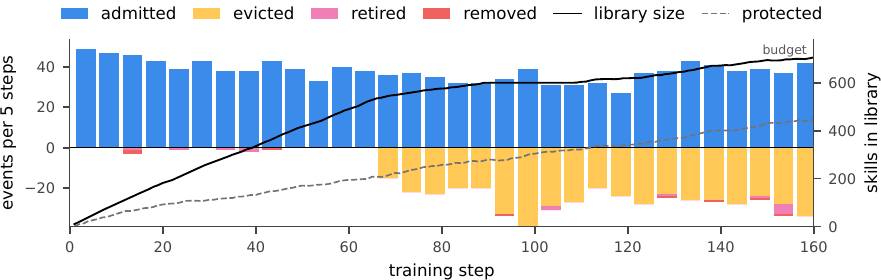}
\caption{\textbf{TrackRL lifecycle events on ALFWorld. }Bars show admissions upward and exits downward over five-step intervals, while lines show active and protected skill counts.}
\label{fig:events}
\end{figure}

\subsection{Skill Library Lifecycle Evolution}
To trace how the library evolves, we aggregate TrackRL's cold-start lifecycle events over five-step intervals in Figure~\ref{fig:events}. Early admissions drive rapid expansion, while deduplication filters repeated proposals. Retirement and harmful removal also begin once the first skills accumulate sufficient evidence, though both remain rare. After the first capacity eviction at step 67, evictions roughly balance admissions and temporarily hold the library near capacity. As protected skills accumulate and no evaluated non-beneficial skills remain eligible for eviction, the soft bound allows the library to exceed capacity rather than evict skills still in incubation. Thus, new skills continue to enter throughout training, while capacity eviction accounts for most exits and retirement and harmful removal occur selectively when supported by evidence.

\subsection{Three Different Skill Lifecycles}
\begin{figure}[!t]
\centering
\includegraphics[width=\linewidth]{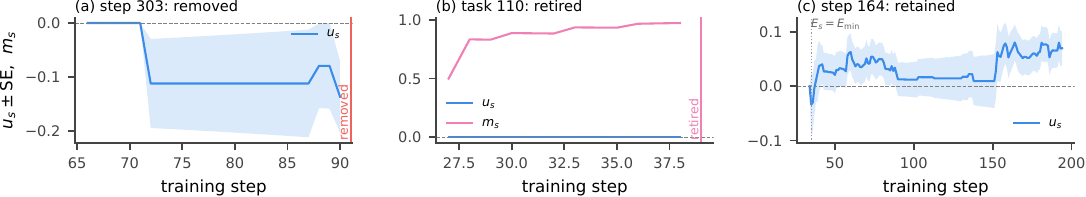}
\caption{\textbf{Three skill lifecycles from ALFWorld.}
The three skills end in removal, retirement, and retention, respectively. Shaded bands indicate utility confidence intervals.}
\label{fig:final_state}
\end{figure}

We trace three skills that are ultimately removed, retired, and retained, respectively (Figure~\ref{fig:final_state}). (1) A step-level skill is protected during incubation and removed at step 91 when its utility reaches $-0.11$ and its upper confidence bound falls below zero. (2) A task-level skill is retired at step 39 as its mastery rises from 0.5 to 0.97 while its utility remains at zero. (3) Another step-level skill remains in the library through training, with utility of 0.07 after 310 comparisons and a confidence interval above zero once estimable. Together, these cases show how skill utility, model mastery, and accumulated evidence guide distinct lifecycle decisions.

\section{Related Work}

\textbf{Skill-Augmented Agentic RL.} Recent skill-augmented agentic RL methods differ in how they use skills. Some distill reusable skills from experience and retrieve them as prompt context for subsequent rollouts and policy optimization \citep{he2026reskill, zhang2026coevolving}. Others adapt the strength, timing, and content of skill guidance to task outcomes or policy competence, sometimes learning when and which skill to invoke \citep{xiao2026uimem, yang2026skillforge, shen2026slim}. A further direction uses skills as privileged training-time context and distills the resulting behavior into a skill-free policy \citep{wang2026skillsd, lu2026skill0, lin2026skillc, zhu2026skill05}. Despite these advances, skill evaluation remains largely based on library-level outcomes, while per-skill estimates rely on one-off validation or post-hoc attribution \citep{wang2026assay, zhao2026coalition}. TrackRL instead identifies each skill's contribution during training and tracks its changing value as the policy evolves, providing evidence for full lifecycle management.

\textbf{Skill Library Lifecycle Management.} Skill libraries require continual maintenance, as stored entries may drift, conflict with subsequent actions, or contain poisoned experience \citep{dabas2026drift, chen2026poisoning}. Existing systems rely on write-time admission \citep{zhang2026consistencygate}, decision-time arbitration \citep{dong2026memarbiter}, pruning, merging, and refactoring \citep{pu2026skillops, lin2026muse}, or consolidation of recurring experience into parameters \citep{xia2026unimem}. These decisions, however, focus on entry properties such as consistency, redundancy, and staleness rather than whether the evolving policy still needs each skill as its capabilities shift \citep{yu2026forget}. As a result, ineffective, internalized, and harmful skills remain difficult to distinguish, although each calls for a different lifecycle decision. TrackRL addresses this distinction through evidence-gated state transitions, using skill utility and model mastery to guide retention, retirement, and removal.

\section{Conclusion}

We presented TrackRL, an agentic RL framework in which skill tracking drives full lifecycle management of the skill library. By turning grouped sampling into an online randomized controlled experiment, TrackRL estimates skill utility, each skill's causal contribution as external scaffolding, and model mastery, the policy's ability to succeed without a corresponding task-level skill. These estimates guide evidence-gated skill lifecycle management from generation and admission through protected incubation to retention, retirement, or removal. TrackRL outperforms baselines in both skill-assisted and skill-free evaluation. Further analyses show how these decisions adapt the library to the evolving policy, closing the loop of policy--library co-evolution.

\bibliography{references}

@article{xia2026skillrl,
  title        = {{SkillRL}: Evolving Agents via Recursive Skill-Augmented Reinforcement Learning},
  author       = {Xia, Peng and Chen, Jianwen and Wang, Hanyang and Liu, Jiaqi and Zeng, Kaide and
                  Wang, Yu and Han, Siwei and Zhou, Yiyang and Zhao, Xujiang and Chen, Haifeng and
                  Zheng, Zeyu and Xie, Cihang and Yao, Huaxiu},
  journal      = {arXiv preprint arXiv:2602.08234},
  year         = {2026},
}

@article{tu2026d2skill,
  title        = {Dynamic Dual-Granularity Skill Bank for Agentic {RL}},
  author       = {Tu, Songjun and Xu, Chengdong and Zhang, Qichao and Zhang, Yaocheng and
                  Lan, Xiangyuan and Li, Linjing and Li, Dong and Zhao, Dongbin},
  journal      = {arXiv preprint arXiv:2603.28716},
  year         = {2026},
}

@article{xiao2026uimem,
  title        = {{UI-Mem}: Self-Evolving Experience Memory for Online Reinforcement Learning in
                  Mobile {GUI} Agents},
  author       = {Xiao, Han and Wang, Guozhi and Wang, Hao and Liu, Shilong and Chai, Yuxiang and
                  Pan, Yue and Zhou, Yufeng and Chen, Xiaoxin and Wen, Yafei and Li, Hongsheng},
  journal      = {arXiv preprint arXiv:2602.05832},
  year         = {2026},
}

@article{yu2026forget,
  title        = {Do Self-Evolving Agents Forget? {C}apability Degradation and Preservation in
                  Lifelong {LLM} Agent Adaptation},
  author       = {Yu, Ye and Yuan, Xiaopeng and Jin, Haibo and Liu, Heming and Yu, Yaoning and
                  Wang, Haohan},
  journal      = {arXiv preprint arXiv:2605.09315},
  year         = {2026},
}

@inproceedings{shridhar2021alfworld,
  title        = {{ALFWorld}: Aligning Text and Embodied Environments for Interactive Learning},
  author       = {Shridhar, Mohit and Yuan, Xingdi and C{\^o}t{\'e}, Marc-Alexandre and
                  Bisk, Yonatan and Trischler, Adam and Hausknecht, Matthew},
  booktitle    = {International Conference on Learning Representations (ICLR)},
  year         = {2021},
  url          = {https://arxiv.org/abs/2010.03768}
}

@inproceedings{yao2022webshop,
  title        = {{WebShop}: Towards Scalable Real-World Web Interaction with Grounded Language
                  Agents},
  author       = {Yao, Shunyu and Chen, Howard and Yang, John and Narasimhan, Karthik},
  booktitle    = {Advances in Neural Information Processing Systems (NeurIPS)},
  year         = {2022},
  url          = {https://arxiv.org/abs/2207.01206}
}

@inproceedings{trivedi2024appworld,
  title        = {{AppWorld}: A Controllable World of Apps and People for Benchmarking Interactive
                  Coding Agents},
  author       = {Trivedi, Harsh and Khot, Tushar and Hartmann, Mareike and Manku, Ruskin and
                  Dong, Vinty and Li, Edward and Gupta, Shashank and Sabharwal, Ashish and
                  Balasubramanian, Niranjan},
  booktitle    = {Proceedings of the 62nd Annual Meeting of the Association for Computational
                  Linguistics (ACL)},
  year         = {2024},
  url          = {https://arxiv.org/abs/2407.18901}
}

@article{pu2026skillops,
  title        = {{SkillOps}: Managing {LLM} Agent Skill Libraries as Self-Maintaining Software
                  Ecosystems},
  author       = {Pu, Hongji and Song, Xinyuan and Zhao, Liang},
  journal      = {arXiv preprint arXiv:2605.13716},
  year         = {2026},
}

@article{liu2026interference,
  title        = {More Is Not Always Better: Cross-Component Interference in {LLM} Agent Scaffolding},
  author       = {Liu, Ming},
  journal      = {arXiv preprint arXiv:2605.05716},
  year         = {2026},
}

@article{wang2023voyager,
  title        = {{VOYAGER}: An Open-Ended Embodied Agent with Large Language Models},
  author       = {Wang, Guanzhi and Xie, Yuqi and Jiang, Yunfan and Mandlekar, Ajay and
                  Xiao, Chaowei and Zhu, Yuke and Fan, Linxi and Anandkumar, Anima},
  journal      = {arXiv preprint arXiv:2305.16291},
  year         = {2023},
}

@article{wang2026skillsd,
  title        = {{Skill-SD}: Skill-Conditioned Self-Distillation for Multi-turn {LLM} Agents},
  author       = {Wang, Hao and Wang, Guozhi and Xiao, Han and Zhou, Yufeng and Pan, Yue and
                  Wang, Jichao and Xu, Ke and Wen, Yafei and Ruan, Xiaohu and Chen, Xiaoxin and
                  Qi, Honggang},
  journal      = {arXiv preprint arXiv:2604.10674},
  year         = {2026},
}

@article{yao2026skillrise,
  title        = {{SkillRise}: Agentic Reinforcement Learning for Cross-Task Skill Evolution},
  author       = {Yao, Zhiyuan and Chen, Yuxin and Lu, Zhengxi and Xu, Zishan and Sun, Yueqing
                  and Guo, Yifu and Lu, Yuquan and Cai, Zhengzhou and Zhang, Kangning and Han,
                  Zhuowen and Wang, Zi-Han and Ye, Ziang and Gu, Qi and Cai, Xunliang and Liu,
                  Weiwen and Shen, Yongliang},
  journal      = {arXiv preprint arXiv:2607.26784},
  year         = {2026},
}

@article{he2026reskill,
  title        = {{ReSkill}: Reconciling Skill Creation with Policy Optimization in Agentic
                  {RL}},
  author       = {He, Zelin and Lin, Haotian and Han, Boran and Zhu, Wei and Fang, Haoyang and
                  Wang, Bernie and Zhu, Xuan and Li, Runze and Reimherr, Matthew},
  journal      = {arXiv preprint arXiv:2606.01619},
  year         = {2026},
}

@article{hu2026persistent,
  title        = {From Interaction Traces to Persistent Skills: Online Evolution for Computer-
                  Use Agents},
  author       = {Hu, Longtao and Liang, Xiao and Zhu, Linchao},
  journal      = {arXiv preprint arXiv:2609.04869},
  year         = {2026},
}

@article{yu2026skill,
  title        = {Skill is Not One-Size-Fits-All: Model-Aware Skill Alignment for {LLM} Agents},
  author       = {Yu, Jianxiang and Zhu, Jiapeng and Lin, Bochen and Cui, Qier and Ding,
                  Zichen and Li, Xiang},
  journal      = {arXiv preprint arXiv:2605.30723},
  year         = {2026},
}

@article{xia2026unimem,
  title        = {{UniMem}: Complementary Episodic-to-Parametric Memory for Boundary-Agnostic
                  Task Streams},
  author       = {Xia, Siyu and Zhang, Chenheng and Wu, Yanting and Li, Haoxuan and Chai,
                  Jiajun and Wang, Xiaohan and Yin, Guojun and Lin, Wei and Lin, Zhouchen and
                  Zhang, Haifeng and Wang, Jun},
  journal      = {arXiv preprint arXiv:2607.26017},
  year         = {2026},
}

@article{dabas2026drift,
  title        = {Memory-Induced Tool-Drift in {LLM} Agents},
  author       = {Dabas, Mahavir and Jeong, Jihyun and Jin, Ming and Jia, Ruoxi},
  journal      = {arXiv preprint arXiv:2605.24941},
  year         = {2026},
}

@article{zhang2026consistencygate,
  title        = {{ConsistencyGate}: Preventing Memory Contamination in {LLM} Agents via Self-
                  Consistency Admission Control},
  author       = {Zhang, Yan and Li, Shibo},
  journal      = {arXiv preprint arXiv:2607.22962},
  year         = {2026},
}

@article{li2026attrimem,
  title        = {{AttriMem}: Attribution-Guided Process Feedback for Agent Memory
                  Construction},
  author       = {Li, Qinfeng and Bao, Yuntai and Yu, Xinyan and Chen, Hongze and Liu, Yanming
                  and Zhu, Huifeng and Jin, Yier and Chen, Jintao and Zhang, Wenqi and Zhang,
                  Xuhong},
  journal      = {arXiv preprint arXiv:2607.21106},
  year         = {2026},
}

@article{wang2026bipace,
  title        = {{BiPACE}: Bisimulation-Guided Policy Optimization with Action Counterfactual
                  Estimation for {LLM} Agents},
  author       = {Wang, Hanyang and Ren, Weijieying and Zhang, Yuxiang and Cao, Ding and Zeng,
                  Zhizhao and Zeng, Ke and Zhao, Tianxiang},
  journal      = {arXiv preprint arXiv:2606.25556},
  year         = {2026},
}

@article{lin2026muse,
  title        = {{MUSE-Autoskill}: Self-Evolving Agents via Skill Creation, Memory,
                  Management, and Evaluation},
  author       = {Lin, Huawei and Li, Peng and Song, Jie and Jiang, Fuxin and Zhang, Tieying},
  journal      = {arXiv preprint arXiv:2605.27366},
  year         = {2026},
}

@article{dong2026memarbiter,
  title        = {{MemArbiter}: Decision-Time Memory Arbitration for Long-Horizon {LLM} Agents},
  author       = {Dong, Jiajun and Hu, Yutao and Fan, Fengrui and Dou, Shihan and Wu, Yueming
                  and Zou, Deqing},
  journal      = {arXiv preprint arXiv:2608.02113},
  year         = {2026},
}

@article{wang2024workflow,
  title        = {Agent Workflow Memory},
  author       = {Wang, Zora Zhiruo and Mao, Jiayuan and Fried, Daniel and Neubig, Graham},
  journal      = {arXiv preprint arXiv:2409.07429},
  year         = {2024},
}

@article{suzgun2025dynamic,
  title        = {Dynamic Cheatsheet: Test-Time Learning with Adaptive Memory},
  author       = {Suzgun, Mirac and Yuksekgonul, Mert and Bianchi, Federico and Jurafsky, Dan
                  and Zou, James},
  journal      = {arXiv preprint arXiv:2504.07952},
  year         = {2025},
}

@article{deng2024explicit,
  title        = {From Explicit {CoT} to Implicit {CoT}: Learning to Internalize {CoT} Step by Step},
  author       = {Deng, Yuntian and Choi, Yejin and Shieber, Stuart},
  journal      = {arXiv preprint arXiv:2405.14838},
  year         = {2024},
}

@article{shao2024deepseekmath,
  title        = {{DeepSeekMath}: Pushing the Limits of Mathematical Reasoning in Open Language Models},
  author       = {Shao, Zhihong and Wang, Peiyi and Zhu, Qihao and Xu, Runxin and Song, Junxiao and
                  Bi, Xiao and Zhang, Haowei and Zhang, Mingchuan and Li, Y. K. and Wu, Y. and Guo, Daya},
  journal      = {arXiv preprint arXiv:2402.03300},
  year         = {2024},
}

@article{zhao2026coalition,
  title        = {Coalition-Aware Skill Reliability for Self-Evolving Agents},
  author       = {Zhao, Qiyan and Zhang, Xiaofeng and Liu, Bo and Chen, Minda and Xiong, Wei and Chen, Jingyang and
                  Ye, Guanting and Yu, Wenhao and Yuan, Xiaosong and Han, Shijie and Wang, Da-Han and Ji, Jianmin and
                  Huang, Fei and Zhang, Xu-Yao},
  journal      = {arXiv preprint arXiv:2608.22610},
  year         = {2026},
}

@article{yang2026skillforge,
  title        = {{SkillForge}: Evolving Verifiable Skills for Reinforcement Learning Agents},
  author       = {Yang, Shidong and Ma, Ziyu and Huang, Tongwen and Wang, Xucong and Li, Renda and Hu, Yiming and
                  Wang, Yong and Chu, Xiangxiang},
  journal      = {arXiv preprint arXiv:2608.24747},
  year         = {2026},
}

@article{chen2026poisoning,
  title        = {When Experience Becomes Instruction: Trajectory Poisoning in Self-Evolving Agent Skill Systems},
  author       = {Chen, Jialuo and Jiang, Lingqi and Deng, Xinhao and Du, Xiaohu and Ma, Jianan and Feng, Yunhao and
                  Qing, Yuqi and Yuan, Zhihao and Du, Linkang and Wang, Jingyi},
  journal      = {arXiv preprint arXiv:2608.05563},
  year         = {2026},
}

@article{auer2002finite,
  title   = {Finite-time Analysis of the Multiarmed Bandit Problem},
  author  = {Auer, Peter and Cesa-Bianchi, Nicol{\`o} and Fischer, Paul},
  journal = {Machine Learning},
  volume  = {47},
  number  = {2-3},
  pages   = {235--256},
  year    = {2002},
  doi     = {10.1023/A:1013689704352}
}

@article{feng2025group,
  title   = {Group-in-Group Policy Optimization for {LLM} Agent Training},
  author  = {Feng, Lang and Xue, Zhenghai and Liu, Tingcong and An, Bo},
  journal = {arXiv preprint arXiv:2505.10978},
  year    = {2025},
}

@article{agashe2024agents,
  title        = {Agent {S}: An Open Agentic Framework that Uses Computers Like a Human},
  author       = {Agashe, Saaket and Han, Jiuzhou and Gan, Shuyu and Yang, Jiachen and
                  Li, Ang and Wang, Xin Eric},
  journal      = {arXiv preprint arXiv:2410.08164},
  year         = {2024},
}

@article{wang2025mobileagente,
  title        = {Mobile-Agent-{E}: Self-Evolving Mobile Assistant for Complex Tasks},
  author       = {Wang, Zhenhailong and Xu, Haiyang and Wang, Junyang and Zhang, Xi and
                  Yan, Ming and Zhang, Ji and Huang, Fei and Ji, Heng},
  journal      = {arXiv preprint arXiv:2501.11733},
  year         = {2025},
}

@article{li2026skillgraph,
  title        = {{SkillGraph}: Skill-Augmented Reinforcement Learning for Agents via Evolving Skill Graphs},
  author       = {Li, Xiaoyuan and Li, Moxin and Bao, Keqin and Ma, Yubo and Wang, Wenjie and Liu, Dayiheng and Feng, Fuli},
  journal      = {arXiv preprint arXiv:2605.12039},
  year         = {2026},
}

@article{shi2026skill1,
  title        = {{Skill1}: Unified Evolution of Skill-Augmented Agents via Reinforcement Learning},
  author       = {Shi, Yaorui and Chen, Yuxin and Lu, Zhengxi and Miao, Yuchun and Liu, Shugui and GU, Qi and Cai, Xunliang and Wang, Xiang and Zhang, An},
  journal      = {arXiv preprint arXiv:2605.06130},
  year         = {2026},
}

@article{lu2026skill0,
  title        = {{SKILL0}: In-Context Agentic Reinforcement Learning for Skill Internalization},
  author       = {Lu, Zhengxi and Yao, Zhiyuan and Wu, Jinyang and Han, Chengcheng and Gu, Qi and Cai, Xunliang and Lu, Weiming and Xiao, Jun and Zhuang, Yueting and Shen, Yongliang},
  journal      = {arXiv preprint arXiv:2604.02268},
  year         = {2026},
}

@article{lin2026skillc,
  title        = {{SkillC}: Learning Autonomous Skill Internalization in {LLM} Agents via Contrastive Credit Assignment},
  author       = {Lin, Hongxiang and Kuai, Zhirui and Xue, Erpeng and Wang, Lei},
  journal      = {arXiv preprint arXiv:2605.27899},
  year         = {2026},
}

@article{zhu2026skill05,
  title        = {{Skill0.5}: Joint Skill Internalization and Utilization for Out-of-Distribution Generalization in Agentic Reinforcement Learning},
  author       = {Zhu, Jiapeng and Yu, Jianxiang and Zhao, Yibo and Han, Chengcheng and Gu, Qi and Cai, Xunliang and Li, Xiang and Qian, Weining},
  journal      = {arXiv preprint arXiv:2605.28424},
  year         = {2026},
}

@article{zhang2026coevolving,
  title        = {Co-Evolving Skill Generation and Policy Optimization},
  author       = {Zhang, Zhiwei and Lin, Yudi and Kuang, Nikki Lijing and Wu, Linlin and Li, Xiaomin and Liu, Songtao and Ma, Fenglong},
  journal      = {arXiv preprint arXiv:2606.08755},
  year         = {2026},
}

@article{shen2026slim,
  title        = {Dynamic Skill Lifecycle Management for Agentic Reinforcement Learning},
  author       = {Shen, Junhao and Zhang, Teng and Zhao, Xiaoyan and Cheng, Hong},
  journal      = {arXiv preprint arXiv:2605.10923},
  year         = {2026},
}

@article{wang2026assay,
  title        = {Not All Skills Help: Measuring and Repairing Agent Knowledge},
  author       = {Wang, Yixuan and Zhou, Yiyang and Liang, Yiming and Zhang, Congyu and Liu, Fuxiao and Zhou, Jiawei and Yao, Huaxiu},
  journal      = {arXiv preprint arXiv:2606.15390},
  year         = {2026},
}
\bibliographystyle{trackrl_arxiv}

\appendix

\clearpage
\setcounter{tocdepth}{2}
\begingroup\renewcommand{\baselinestretch}{1.15}\selectfont
\tableofcontents
\endgroup
\clearpage

{\centering\bfseries\Large Appendix\par}

The appendix provides the experimental details, implementation, prompts, and supplementary results referenced in the main text. Appendix~\ref{app:setup} describes the two benchmarks, their splits, and the metrics we report. Appendix~\ref{app:baselines} describes each baseline and the settings shared across methods. Appendix~\ref{app:evaluation} gives the evaluation protocol, including the two evaluation modes, the evaluation prompts, and the episode budget. Appendix~\ref{app:impl} gives the implementation of TrackRL, covering the skill record with an example of each granularity, the skill pools and retrieval, the full training step as an algorithm, the hyperparameters, and the computational cost. Appendix~\ref{app:prompts} lists the prompt used for skill generation and the format in which skills are injected. Appendix~\ref{app:results} reports additional results, namely the per-class breakdown on the unseen split of ALFWorld, internalization by task family, and the library under warm start. Appendix~\ref{app:impacts} discusses the broader impacts of this work.

\section{Benchmarks}
\label{app:setup}
\label{app:benchmarks}

We evaluate on two benchmarks of different form, a text-based household environment and a code-based environment of applications and APIs.

\textbf{ALFWorld} \citep{shridhar2021alfworld} is a text-based household environment with six task families, namely Pick, Clean, Cool, Look, Heat, and Pick2. We train on the official training games and evaluate on the \emph{seen} split of 140 games and on the \emph{unseen} split of 134 games in room layouts absent from training. We follow the standard protocol of RL-trained ALFWorld agents \citep{feng2025group, xia2026skillrl}, with 128 games sampled per split, sampling temperature 0.4, at most 50 turns, and a frozen library during evaluation, and we report the success rate (SR).

\textbf{AppWorld} \citep{trivedi2024appworld} is a code-based environment of 9 applications and 457 APIs in which the agent completes tasks by writing API calls. We train on the train split and evaluate on \emph{test-normal} with 168 tasks drawn from 56 scenarios and on \emph{test-challenge} with 417 tasks drawn from 139 scenarios, using the official ReAct prompt, at most 40 turns, and greedy decoding. A scenario is a single situation instantiated as three related task variants, so both splits contain exactly three tasks per scenario. We report task goal completion (TGC), which counts a task as solved when all of its unit tests pass, and scenario goal completion (SGC), which counts a scenario as solved only when all three of its tasks are solved, making SGC the stricter of the two metrics.

\section{Baselines}
\label{app:baselines}

We compare TrackRL against the three groups of methods used in Section~\ref{sec:exp}. Every method sees the same environments, the same interaction prompt, and the same turn cap, and every trained method additionally shares the policy model, the group size, and the batch size, so that a difference between two rows is attributable to the method rather than to its setting.

\textbf{Proprietary LLMs.} Two closed-source models under direct prompting, with no training and no skill library, are included to show what a base model achieves on these tasks rather than as trained competitors.
\begin{itemize}
\item \textbf{Gemini-2.5-Flash} is prompted directly on the same tasks, with the same interaction prompt and the same turn cap as every trained method.
\item \textbf{GPT-5.6-Luna} follows the same direct-prompting protocol. Both models also serve as reflection models for the skill-augmented methods, so the comparison also shows what the model generating the skills can do on the task itself.
\end{itemize}

\textbf{Skill-free Models.} Evaluating the policy without a skill library measures what it can accomplish using its learned capabilities alone. This distinguishes improvements retained by the policy from gains that depend on retrieving skills as external support at evaluation time.
\begin{itemize}
\item \textbf{Origin} is the untrained base model, Qwen3-4B-Instruct-2507, prompted with the same interaction format and given no library. It provides a common reference for measuring gains from training.
\item \textbf{SFT} is the base model after supervised fine-tuning on demonstration trajectories, with no reinforcement learning and no library. The SFT model appears in our tables both as a standalone baseline and as the second initialization for every RL method.
\item \textbf{GRPO} \citep{shao2024deepseekmath} is group relative policy optimization, which removes the value network and standardizes the reward of each episode against the other episodes of its group, so that the group itself supplies the baseline. GRPO underlies TrackRL's policy optimization but uses no skill mechanism, so its performance gap with TrackRL reflects the contribution of the skill library under the same training objective.
\end{itemize}

\textbf{Skill-augmented RL methods.} These methods build a skill library during training and inject retrieved skills into the prompt, as TrackRL does. Both add an intrinsic reward that favors skill-conditioned over skill-free episodes. We omit it in our runs to avoid rewarding reliance on the skill library as external scaffolding rather than task completion.
\begin{itemize}
\item \textbf{SkillRL} \citep{xia2026skillrl} is recursive skill-augmented reinforcement learning, in which the agent distills skills from its own trajectories and reuses them in later episodes as the library accumulates. Skills enter the library and are never removed or retired, so SkillRL is our reference point for what unmanaged accumulation costs over a long run.
\item \textbf{D2Skill} \citep{tu2026d2skill} is a dynamic dual-granularity skill bank that maintains task-level and step-level skills, the same two granularities we adopt, and updates them during training. It is the closest method to ours and the most informative comparison, because it differs mainly in what it measures. D2Skill scores the retrieved bundle rather than the individual skills inside it, so a skill can inherit credit from whatever was retrieved alongside it, which is the free-riding that per-skill randomization is designed to remove.
\end{itemize}

We train every skill-augmented RL method from both the base model and the SFT checkpoint. We also run every skill-augmented method with each of two reflection models, GPT-5.6-Luna and Gemini-2.5-Flash, so that the findings do not depend on a single reflection model. All skill-augmented methods use the same retrieval embedding model, holding this component fixed when comparing how they manage their libraries.

\section{Evaluation}
\label{app:evaluation}
\label{app:eval_prompts}

\textbf{Evaluation modes.} Each checkpoint is evaluated in two modes on the same tasks. In the \emph{skill-assisted} mode, skills are retrieved and injected exactly as in training, which measures the agent and its library as a whole and matches the reporting of prior work. In the \emph{skill-free} mode, nothing is retrieved and the prompt is identical to that of plain GRPO, which measures the capability that has entered the policy parameters. The difference between the two modes on the same checkpoint is the net effect of the library on that policy, and for a method without a library the two modes coincide.

\textbf{Evaluation prompts.} Evaluation uses the same interaction prompt as training. On ALFWorld this is the standard ReAct-style prompt of the environment, with the injected block inserted exactly as in Appendix~\ref{app:injection_format}, and on AppWorld it is the official ReAct prompt distributed with the benchmark. Two things differ from training. The library is frozen, so no skill is admitted, retired, or removed during an evaluation pass, and nothing is randomized, so every selected skill is injected in the skill-assisted mode and the block is absent in the skill-free mode. The skill-free prompt is therefore identical to the prompt of plain GRPO, which is what makes the two modes comparable across methods with and without a library.

\textbf{Episode budget.} An episode ends when the task is solved, when the environment declares failure, or when the turn cap is reached. The cap is 50 turns on ALFWorld and 40 turns on AppWorld, and it is the same for every method, including the proprietary models under direct prompting. The cap matters for the reported numbers because an unsolved episode consumes the whole budget. Across the ALFWorld validation passes, failed episodes average close to the 50-turn cap while solved episodes finish in well under half of it, so a method with a lower success rate also consumes more interaction per solved task.

\section{Implementation Details}
\label{app:impl}

\subsection{Skills Organization, Retrieval, and Injection}
\label{app:skills}
\label{app:retrieval}

\textbf{Skill record.} A skill is a short textual record with three fields. The \emph{title} is a few words naming the skill and appears as the heading of its entry in the injected context. The \emph{principle} is one or two sentences stating what to do and carries the substantive content of the skill. The \emph{when-to-apply} clause is one sentence describing the situation the skill addresses. This clause serves as the retrieval key $\kappa(s)$ in Eq.~(\ref{eq:retrieve}), so a skill is matched by the situation it claims to cover rather than by the wording of its advice. Alongside its text, a skill stores its tracked statistics, its exposure count $N_s$, and the step at which it was admitted. A task-level skill additionally records the task family of the trajectory from which it was distilled, which supplies $f(s)$ in Eq.~(\ref{eq:skill_retirement}).

\textbf{Skill Pools.} The two pools differ in when they are queried and how long injected skills remain in context. The task-level pool $\mathcal{P}_{\mathrm{task}}$ is queried once at the beginning of each episode using the task description. When injected, a task-level skill appears before the first observation and remains in context throughout the episode. Retrieval is restricted to the episode's task family, with no backfilling from other families. The step-level pool $\mathcal{P}_{\mathrm{step}}$ is queried at every turn using the task description concatenated with the current observation. When injected, a step-level skill follows that observation and guides only the next action. Both pools start empty and are populated solely through reflection on failures during training. 

The example below shows a task-level and a step-level skill generated by the reflection model for the heating family of ALFWorld, written as the JSON records returned by reflection (Appendix~\ref{app:reflection_prompt}).

\begin{promptbox}
\begin{verbatim}
Task-level skill
{
  "title": "Complete the heating-and-storage sequence",
  "principle": "Locate the potato, place it in the microwave,
    activate the microwave to heat it, then remove it and put it
    in the fridge; do not confuse opening or closing the appliance
    with heating.",
  "when_to_apply": "When the task requires heating an object and
    subsequently storing it in a different receptacle"
}

Step-level skill
{
  "title": "Search likely cooking locations systematically",
  "principle": "After checking the fridge, inspect the stove or
    other likely egg locations before wandering through unrelated
    containers.",
  "when_to_apply": "When the target object has not been found and
    the task involves heating or cooking it"
}
\end{verbatim}
\end{promptbox}

\textbf{Retrieval and Assignment.} We compare queries and retrieval keys by the cosine similarity of their embeddings. If no key exceeds the retrieval threshold, we return an empty set rather than a weak match. Exposure counts used in the reranking bonus accumulate without discounting. We draw each task-level assignment once at episode onset and each step-level assignment upon first retrieval, reusing it whenever the skill is retrieved again in the same episode. In the two skill-free episodes reserved per group, retrieval and reranking remain unchanged, but the corresponding assignments are fixed to zero, so their outcomes enter the control arm for every skill selected there. Injected skills appear in up to two short Markdown sections, one per granularity (Appendix~\ref{app:injection_format}). When no skill is injected, we omit the entire block, leaving the same prompt as in plain GRPO.

\subsection{Training Procedure}
\label{app:algorithm}
\label{app:estimators}
\label{app:lifecycle_detail}

This section states one full training step of TrackRL with every equation of Section~\ref{sec:tracking} and Section~\ref{sec:lifecycle} written out in the place where it is applied. Table~\ref{tab:notation} collects the notation, and Algorithm~\ref{alg:trackrl} gives the procedure. 

\begin{table}[h]
\centering
\footnotesize
\caption{Mathematical Notation. The upper block is the sampling and optimization setting, the middle block is the state TrackRL tracks for a skill, and the lower block is the parameters that govern retrieval and the lifecycle.}
\label{tab:notation}
\begin{tabular}{lll}
\toprule
\textbf{Symbol} & \textbf{Meaning} & \textbf{Introduced in} \\
\midrule
$\pi_\theta,\ \pi_{\mathrm{old}},\ \pi_{\mathrm{ref}}$ & current, behavior, and reference policy & Eq.~(\ref{eq:grpo}) \\
$M,\ B$ & tasks per step and episodes per group & Section~\ref{sec:tracking} \\
$g,\ e$ & a group of episodes and a single episode & Section~\ref{sec:tracking} \\
$x,\ \kappa(s)$ & retrieval query and the retrieval key of skill $s$ & Eq.~(\ref{eq:retrieve}) \\
$y_e,\ \phi_e,\ R_e,\ A_e$ & outcome, invalid-action fraction, reward, group-normalized advantage & Eq.~(\ref{eq:grpo}) \\
$z_{s,e}$ & injection assignment of skill $s$ in episode $e$ & Eq.~(\ref{eq:randomized_injection}) \\
$\mathcal{P}_{\mathrm{task}},\ \mathcal{P}_{\mathrm{step}},\ \mathcal{H}$ & task-level pool, step-level pool, record of harmful skills & Section~\ref{sec:lifecycle} \\
\midrule
$n^{+}_{s,g},\ n^{-}_{s,g}$ & treatment and control arm sizes of $s$ in group $g$ & Eq.~(\ref{eq:diff}) \\
$c^{+}_{s,g},\ c^{-}_{s,g}$ & successes in the two arms & Eq.~(\ref{eq:diff}) \\
$d_{s,g},\ w_{s,g}$ & paired comparison and its reliability weight & Eq.~(\ref{eq:diff}) \\
$W_s,\ S_s,\ Q_s,\ G_s$ & discounted running statistics of $s$ & Eq.~(\ref{eq:suffstat}) \\
$u_s,\ \mathrm{SE}(u_s)$ & utility estimate and its standard error & Eq.~(\ref{eq:utility}) \\
$n^{-}_{s},\ c^{-}_{s}$ & discounted control-arm episode and success counts & Eq.~(\ref{eq:armcounts}) \\
$m_s,\ \mathrm{SE}(m_s)$ & mastery estimate and its standard error & Eq.~(\ref{eq:mastery}) \\
$E_s$ & undiscounted evidence count since admission & Section~\ref{sec:tracking} \\
$N_s,\ N_{\mathcal{P}}$ & exposure count of $s$ and of its pool & Eq.~(\ref{eq:score}) \\
$b_{f(s)}$ & skill-free success rate of the family of $s$ & Eq.~(\ref{eq:skill_retirement}) \\
\midrule
$\delta,\ m,\ K$ & similarity threshold, candidates kept, skills selected & Eq.~(\ref{eq:retrieve}) \\
$\alpha,\ \eta$ & relevance-evidence balance and exploration coefficient & Eq.~(\ref{eq:score}) \\
$p$ & injection probability in randomized episodes & Eq.~(\ref{eq:randomized_injection}) \\
$\gamma$ & recency discount & Eq.~(\ref{eq:suffstat}) \\
$\lambda,\ E_{\min},\ m_{\min}$ & confidence level, evidence threshold, mastery floor & Eq.~(\ref{eq:skill_removal}), Eq.~(\ref{eq:skill_retirement}) \\
$\delta_{\mathrm{dup}},\ A,\ C$ & deduplication threshold, admission cap, soft capacity & Eq.~(\ref{eq:skill_admission}), Section~\ref{sec:lifecycle} \\
$\epsilon,\ \beta,\ \nu$ & clipping range, KL weight, invalid-action penalty & Eq.~(\ref{eq:grpo}) \\
\bottomrule
\end{tabular}
\end{table}
\clearpage
\begin{algorithm}[H]
\caption{One training step of TrackRL. The library is frozen for the duration of the step.}
\label{alg:trackrl}
\small
\begin{algorithmic}[1]
\REQUIRE $\pi_\theta$, pools $\mathcal{P}_{\mathrm{task}}, \mathcal{P}_{\mathrm{step}}$, $\mathcal{H}$, per-skill state, family baselines $b_f$
\STATE \textbf{Stage 1. Sampling.} Sample $M$ tasks and roll out one group $g$ of $B$ episodes each under $\pi_{\mathrm{old}}$.
\FOR{each group $g$}
    \STATE Reserve one task-level and one step-level skill-free episode, which use $p = 0$.
    \FOR{each episode $e \in g$, per episode for task-level and per turn for step-level}
        \STATE $\mathcal{R}(x;\mathcal{P}) = \operatorname{Top}_{m}\big\{ s \in \mathcal{P} \mid \operatorname{sim}(x, \kappa(s)) \ge \delta \big\}$ \hfill Eq.~(\ref{eq:retrieve})
        \STATE $\operatorname{score}(s) = \alpha\, \widetilde{\operatorname{sim}}(x, \kappa(s)) + (1-\alpha)\big[ u_s + \eta \sqrt{\log(1+N_{\mathcal{P}}) / (1+N_s)} \big]$ \hfill Eq.~(\ref{eq:score})
        \STATE Take the top $K$, draw $z_{s,e} \sim \operatorname{Bernoulli}(p)$ independently per skill \hfill Eq.~(\ref{eq:randomized_injection})
        \STATE Inject the skills with $z_{s,e} = 1$, act, and increment $N_s$ and $N_{\mathcal{P}}$
    \ENDFOR
    \STATE Observe $y_e$ and record each episode with its offered skills and assignments
\ENDFOR
\STATE \textbf{Stage 2. Tracking.} Let $\mathcal{G}_s$ be the groups of this step in which both arms of $s$ are non-empty.
\FOR{each skill $s$ and each $g \in \mathcal{G}_s$}
    \STATE $d_{s,g} = \dfrac{c^{+}_{s,g}}{n^{+}_{s,g}} - \dfrac{c^{-}_{s,g}}{n^{-}_{s,g}}$, \quad $w_{s,g} = \dfrac{n^{+}_{s,g} n^{-}_{s,g}}{n^{+}_{s,g} + n^{-}_{s,g}}$ \hfill Eq.~(\ref{eq:diff})
\ENDFOR
\FOR{each skill $s$ in either pool}
    \STATE $W_s \leftarrow \gamma W_s + \sum_{g} w_{s,g}$, \quad $S_s \leftarrow \gamma S_s + \sum_{g} w_{s,g} d_{s,g}$, \quad $Q_s \leftarrow \gamma Q_s + \sum_{g} w_{s,g} d_{s,g}^{2}$, \quad $G_s \leftarrow \gamma G_s + |\mathcal{G}_s|$ \hfill Eq.~(\ref{eq:suffstat})
    \STATE $u_s = S_s / W_s$, \quad $\mathrm{SE}(u_s) = \sqrt{(Q_s - S_s^{2}/W_s) / (W_s (G_s - 1))}$ for $G_s \ge 2$, \quad $E_s \leftarrow E_s + |\mathcal{G}_s|$ undiscounted \hfill Eq.~(\ref{eq:utility})
    \IF{$s$ is task-level}
        \STATE $n^{-}_{s} \leftarrow \gamma n^{-}_{s} + \sum_{g} n^{-}_{s,g}$, \quad $c^{-}_{s} \leftarrow \gamma c^{-}_{s} + \sum_{g} c^{-}_{s,g}$ \hfill Eq.~(\ref{eq:armcounts})
        \STATE $m_s = \dfrac{c^{-}_{s} + 1}{n^{-}_{s} + 2}$, \quad $\mathrm{SE}(m_s) = \sqrt{m_s (1 - m_s) / (n^{-}_{s} + 3)}$ \hfill Eq.~(\ref{eq:mastery})
    \ENDIF
\ENDFOR
\STATE Update each family baseline $b_f$ by the same recursion over the skill-free episodes of family $f$
\STATE \textbf{Stage 3. Lifecycle.}
\FOR{each skill $s$ in either pool}
    \IF{$E_s \ge E_{\min}$ \AND $u_s + \lambda\, \mathrm{SE}(u_s) < 0$}
        \STATE Remove $s$, \quad $\mathcal{H} \leftarrow \mathcal{H} \cup \{s\}$ \hfill Eq.~(\ref{eq:skill_removal})
    \ELSIF{$s$ task-level \AND $E_s \ge E_{\min}$ \AND $u_s \le 0$ \AND $m_s - \lambda\, \mathrm{SE}(m_s) \ge \max\{ b_{f(s)}, m_{\min} \}$}
        \STATE Retire $s$ \hfill Eq.~(\ref{eq:skill_retirement})
    \ENDIF
\ENDFOR
\FOR{each candidate from reflection on the lowest-scoring groups, at most $A$ per pool}
    \IF{$\max_{s \in \mathcal{P} \cup \mathcal{H}} \operatorname{sim}(s_{\mathrm{new}}, s) < \delta_{\mathrm{dup}}$}
        \STATE Admit $s_{\mathrm{new}}$ with empty statistics \hfill Eq.~(\ref{eq:skill_admission})
    \ENDIF
\ENDFOR
\WHILE{$|\mathcal{P}| > C$ \AND $\{ s : E_s \ge E_{\min} \wedge u_s \le 0 \} \neq \emptyset$}
    \STATE Evict the eligible skill of smallest $u_s$
\ENDWHILE
\STATE \textbf{Stage 4. Policy update.} $R_e = y_e - \nu \phi_e$, \quad $A_e = (R_e - \operatorname{mean}_{e' \in g} R_{e'}) / \operatorname{std}_{e' \in g} R_{e'}$
\STATE $\max_\theta\ \mathcal{J}(\theta) = \mathbb{E}_{g}\big[ \tfrac{1}{B} \sum_{e \in g} \tfrac{1}{L_e} \sum_{k} \min( \rho_{e,k} A_e, \operatorname{clip}(\rho_{e,k}, 1{\pm}\epsilon) A_e ) \big] - \beta D_{\mathrm{KL}}(\pi_\theta \| \pi_{\mathrm{ref}})$ \hfill Eq.~(\ref{eq:grpo})
\end{algorithmic}
\end{algorithm}
\newpage

\subsection{Hyperparameters}
\label{app:hparams}

Table~\ref{tab:hp_rl} lists the reinforcement learning hyperparameters, which are shared by every trained method so that the comparison isolates the library rather than the optimizer. Table~\ref{tab:hp_track} lists the parameters of TrackRL together with the symbol each one carries in Section~\ref{sec:tracking} and Section~\ref{sec:lifecycle}.

\begin{table}[h]
\centering
\footnotesize
\caption{Reinforcement learning hyperparameters, shared by all trained methods.}
\label{tab:hp_rl}
\begin{tabular}{llc}
\toprule
\textbf{Parameter} & \textbf{Symbol} & \textbf{Value} \\
\midrule
Policy model & $\pi_\theta$ & Qwen3-4B-Instruct-2507 \\
Tasks per training step & $M$ & 16 \\
Episodes per group & $B$ & 8 \\
Episodes per training step & $M \cdot B$ & 128 \\
Training steps & --- & 160 \\
Learning rate & --- & $1 \times 10^{-6}$ \\
Mini-batch size (policy update) & --- & 128 \\
Clipping range & $\epsilon$ & 0.2 \\
KL regularization weight & $\beta$ & 0.01 \\
Invalid-action penalty & $\nu$ & 1.0 \\
Maximum prompt length & --- & 4096 tokens \\
Maximum response length per turn & --- & 512 tokens \\
Rollout engine & --- & vLLM \\
Validation games per evaluation & --- & 128 \\
Validation sampling temperature (ALFWorld) & --- & 0.4 \\
Maximum turns (ALFWorld / AppWorld) & --- & 50 / 40 \\
\bottomrule
\end{tabular}
\end{table}

\begin{table}[h]
\centering
\footnotesize
\caption{TrackRL parameters. The upper block is introduced by skill tracking and lifecycle management. The lower block is the retrieval configuration shared with the skill-augmented baselines.}
\label{tab:hp_track}
\begin{tabular}{llc}
\toprule
\textbf{Parameter} & \textbf{Symbol} & \textbf{Value} \\
\midrule
Injection probability, randomized episodes & $p$ & $1/2$ \\
Skill-free episodes reserved per group & --- & 1 task-level, 1 step-level \\
Recency discount & $\gamma$ & 0.98 \\
Evidence threshold for any exit & $E_{\min}$ & 10 \\
Confidence level of both one-sided tests & $\lambda$ & 2 \\
Absolute mastery floor for retirement & $m_{\min}$ & 0.5 \\
Soft capacity per pool & $C$ & 300 \\
Admission cap per pool per step & $A$ & 5 \\
Deduplication threshold & $\delta_{\mathrm{dup}}$ & 0.90 \\
Exploration coefficient in reranking & $\eta$ & 0.2 \\
Relevance-evidence balance in reranking & $\alpha$ & 0.1 \\
\midrule
Embedding model & --- & Qwen3-Embedding-0.6B \\
Retrieval similarity threshold & $\delta$ & 0.70 \\
Candidates retrieved per pool & $m$ & 10 \\
Skills selected per pool & $K$ & 3 \\
Skill pools & $\mathcal{P}_{\mathrm{task}}, \mathcal{P}_{\mathrm{step}}$ & task-level, step-level \\
Reflection model & --- & GPT-5.6-Luna or Gemini-2.5-Flash \\
\bottomrule
\end{tabular}
\end{table}

\subsection{Computational Resources}
\label{app:compute}

Each run trains a 4B policy on a single node of high-performance GPUs, with vLLM serving the rollouts. On ALFWorld, for example, a training step collects 128 episodes of at most 50 turns, and each turn requires one generation call from the policy, so a step performs up to 6{,}400 such calls. Its wall-clock cost is therefore dominated by autoregressive generation rather than by the environment or the policy update. The turns of an episode are sequential, so these calls form up to fifty successive rounds of vLLM inference, each batched across the episodes of the step. The ALFWorld simulator is text-based and steps in negligible time, and the gradient update accounts for a small share of the step. Validation runs every five steps in both evaluation modes and on both splits. A single validation pass over 128 ALFWorld games costs roughly as much as a training step, making it the largest fixed overhead of the protocol. Every method incurs this cost.
The components TrackRL adds are computationally inexpensive. Tracking uses the episode outcomes GRPO already collects and updates a constant number of scalars per skill per step. Lifecycle management performs threshold comparisons over the pool and computes one embedding per candidate for the admission test. Retrieval makes one embedding call per query and reranks at most ten candidates. Reflection makes one call to an external model per failed trajectory selected at that step. This cost is not negligible, but it is shared with the skill-augmented baselines. TrackRL adds no rollouts. The treatment and control arms of Eq.~(\ref{eq:randomized_injection}) are formed from the $B$ episodes GRPO already samples, so a TrackRL step and a GRPO step collect the same number of episodes.

\section{Prompt Templates}
\label{app:prompts}

\subsection{Skill Generation and Reflection}
\label{app:reflection_prompt}

At each training step, the groups with the lowest success rates supply their failed trajectories to the reflection model. Each call receives one failed trajectory, and when the same group contains a solved episode of the same task, that successful trajectory is supplied as a reference. The model is asked for three outputs. The \emph{first error step} identifies the turn at which the failed trajectory first went wrong and anchors the step-level skill. The \emph{step reflection} is one step-level skill addressing that turn, and the \emph{task reflection} is one task-level skill for the task as a whole. The two skills are returned as JSON objects with exactly the three fields of Appendix~\ref{app:skills}, allowing one record to serve as both the retrieval key and the injected text.

\begin{promptbox}
\begin{verbatim}
You are given one failed trajectory and one successful trajectory for
the same task. Analyze them and produce exactly three outputs.

Task: {task description}
Task Type: {task family}

Failed trajectory:
{turn-by-turn observations and actions}

Successful trajectory (reference):
{turn-by-turn observations and actions}

Output the following in order (use the exact section headers):

1) FIRST_ERROR_STEP: N
   Where N is the 1-based turn number in the *failed* trajectory where
   the agent first went wrong (e.g. Turn 1, Turn 2). Use 0 if unclear.

2) STEP_REFLECTION (one step-level experience for that step only):
   Output a JSON object with exactly: "title", "principle",
   "when_to_apply". This should be a concise experience for what to do
   at that specific step/situation (e.g. "At this step you should
   ...").

3) TASK_REFLECTION (one task-level skill for the whole task):
   Output a JSON object with exactly: "title", "principle",
   "when_to_apply". This summarizes for the *whole task*: what to
   avoid and how to succeed in this kind of task (will be retrieved by
   task description).

Output format (use these exact labels):
FIRST_ERROR_STEP: N

STEP_REFLECTION:
<single JSON object>

TASK_REFLECTION:
<single JSON object>
\end{verbatim}
\end{promptbox}

When no successful trajectory is available for the task, the first paragraph and the reference trajectory are omitted and the model is given the failed trajectory alone. A candidate whose JSON cannot be parsed is discarded. Surviving candidates enter the admission test of Eq.~(\ref{eq:skill_admission}), and at most $A = 5$ per pool are admitted per step.

\subsection{Retrieval and Injection Format}
\label{app:injection_format}

The skills with $z_{s,e}=1$ are rendered in up to two Markdown sections and placed in the prompt as described in Appendix~\ref{app:retrieval}. Only the title and principle are shown. The when-to-apply clause serves as the retrieval key and is not injected, so the context the policy reads states what to do rather than restating the situation it already observes. The example below uses skills from the cold-start library at step 160.

\begin{promptbox}
\begin{verbatim}
### Task-level experience (for this kind of task)
- **Complete required object-state changes before placement**: For
  tasks requiring a clean object in a receptacle, locate and take the
  object, clean it using an available sink, then place it in the
  target cabinet.

### Step-level experience (relevant to current situation)
- **Search likely surfaces systematically**: After opening the fridge
  and not finding the target, inspect nearby cabinets and countertops
  rather than going to an irrelevant dining table.
\end{verbatim}
\end{promptbox}

A section is omitted when no skill of that granularity is injected, and the whole block is omitted when neither pool contributes. The latter is the case in the reserved skill-free episodes and at the start of training when both pools are still empty. The task-level section is written once before the first observation and stays in context for the episode, whereas the step-level section is rewritten after each observation.

\section{Additional Results}
\label{app:results}

\subsection{Per-Class Breakdown on ALFWorld}
\label{app:perclass}

Table~\ref{tab:alfworld} gives the per-class results on the seen split. Table~\ref{tab:unseen_perclass} gives the corresponding breakdown on the unseen split, for the runs we trained ourselves and whose per-episode validation records we retain.

\begin{table}[h]
\centering
\footnotesize
\setlength{\tabcolsep}{4pt}

\begin{tabular}{llcccccc|c}
\toprule
\textbf{Method} & \textbf{Mode} & Pick & Clean & Cool & Look & Heat & Pick2 & \textbf{Overall} \\
\midrule
D2Skill (Luna) & skill-assisted & 46.7 & 63.0 & 47.8 & 45.5 & 46.7 & 72.7 & 54.7 \\
 & skill-free & 55.9 & 40.0 & 70.8 & 54.5 & 35.7 & 45.0 & 52.3 \\
\addlinespace[2pt]
\textbf{TrackRL} (Luna) & skill-assisted & 61.3 & 65.2 & 82.6 & 61.5 & 92.9 & 75.0 & 71.9 \\
 & skill-free & 63.3 & 66.7 & 52.2 & 91.7 & 75.0 & 52.2 & 64.1 \\
\addlinespace[2pt]
\textbf{TrackRL} (Luna) + SFT & skill-assisted & 100.0 & 96.2 & 96.0 & 100.0 & 100.0 & 95.2 & 97.7 \\
 & skill-free & 87.5 & 88.0 & 100.0 & 100.0 & 100.0 & 100.0 & 95.3 \\
\bottomrule
\end{tabular}
\caption{\textbf{Results on the unseen split of ALFWorld by task class.} SR (\%) by task class, with the overall values of Table~\ref{tab:alfworld}.}
\label{tab:unseen_perclass}
\end{table}

Under SFT initialization the two modes nearly coincide on the unseen split, at 97.7\% skill-assisted against 95.3\% skill-free. Once the policy is sufficiently strong the library neither adds to nor subtracts from its unseen performance, which is the regime in which retirement rather than retention is the common outcome.

\subsection{Internalization by Task Family}
\label{app:family}

\begin{figure}[!t]
\centering
\includegraphics[width=\linewidth]{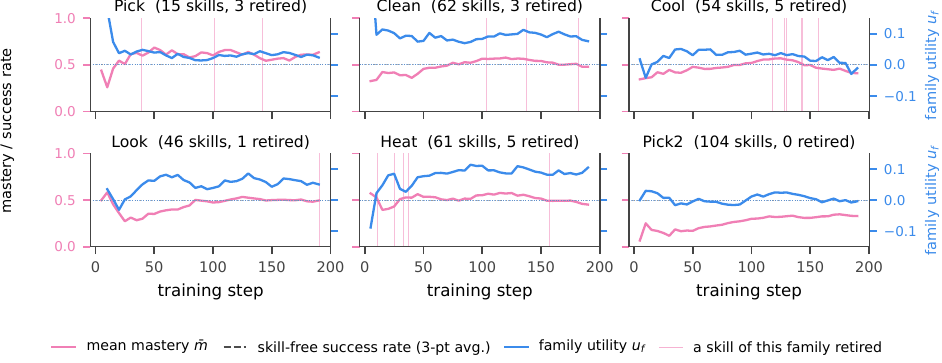}
\caption{\textbf{Task-level skills by family (ALFWorld, base model, Luna reflection).} One panel per family, showing the mean mastery of the family's active task-level skills on the left axis, the success rate of the policy without the library on the seen validation games of that family on the same axis as a dashed line and as a three-point moving average since each family has about twenty games per validation, the family utility pooled over the family's skills on the right axis, and one vertical tick per retirement of a skill of that family. Panel titles give the number of active task-level skills of the family at the last step and the number retired.}
\label{fig:family}
\end{figure}

Figure~\ref{fig:family} tracks three quantities for each of the six ALFWorld task families: the family utility of its task-level skills, their mean mastery, and the policy's success rate without the library. Family utility pools the within-group utility estimates of Section~\ref{sec:tracking} across comparisons of the family's skills, measuring what those skills add to an episode from that family at each step. The three quantities evolve as the lifecycle framework anticipates, with the six families showing three distinct patterns.

\textbf{Internalized families.} In Pick, the policy solves 81\% of validation games without skills by the end of training, up from 66\% at the start. Family utility falls from 0.09 to 0.04, and three skills are retired. In Cool, utility peaks at 0.05 around steps 60 to 120, when mastery is highest. Five skills are retired between steps 115 and 160, and utility returns to zero by the end.

\textbf{Families that continue to require support.} Clean maintains a family utility of 0.10 throughout training, with skill-free success between 21\% and 32\%. Heat and Look end at 0.09 and 0.06, respectively. Where the policy continues to fail without skills, family utility persists.

\textbf{The family the library does not affect.} In Pick2, the hardest family, utility remains at zero, mastery stays at 0.33, and no skill is retired. The policy improves from 16\% to 39\% through reinforcement learning alone, while the library neither improves nor degrades its performance.

\subsection{Library Size for the SFT-Initialized Policy}
\label{app:libsize_sft}

The bottom row of Figure~\ref{fig:library_size} tracks the library of the SFT-initialized policy. Its trajectory differs from the cold-start run more in magnitude than in shape. The library grows rapidly to 135 skills in the first 40 steps, then levels off, reaching 212 skills at step 160, including 80 task-level and 132 step-level skills. Neither pool approaches the soft capacity of $C=300$, so the capacity rule never binds and every exit in this run is evidence-gated rather than capacity-based.

The library remains small because skills are generated only from failures. The SFT-initialized policy already solves most training tasks, so few groups qualify among the lowest-scoring groups at each step, few candidates are proposed, and the admission cap of $A=5$ per pool rarely binds. The composition of exits also differs. Under SFT initialization, 33 task-level skills are retired and 9 skills are deleted as harmful or duplicated by step 160. Retirement thus accounts for most exits and covers roughly a third of all task-level skills that were ever active. Under base-model initialization, the proportions are reversed, with 14 retirements against 497 deletions, and retirement covers about 3\% of task-level skills that were ever active. The same lifecycle rules therefore produce a small, heavily internalized library under a strong initialization and a large, capacity-bound library under a weak one, without changing the thresholds between runs.

\section{Broader Impacts}
\label{app:impacts}
The components this work adds are a bookkeeping layer over an existing training pipeline, and their main external effect is to make an agent's accumulated context trackable. Every estimate and every lifecycle transition is recorded together with the statistics that justified them, so a third party can replay offline why a skill was kept, retired, or removed. We regard this as the useful direction of the contribution, since a skill library is otherwise an opaque store of model-generated text that silently shapes an agent's behavior.

The same mechanism carries the corresponding risk. Skills are distilled by a language model from its own failures, so they can encode mistaken or unsafe generalizations, and a skill that is merely useless rather than harmful is retained as long as its measured utility is not negative. The evidence threshold delays every exit by design, so a harmful skill remains in the library until enough comparisons accumulate. Deployment beyond simulated benchmarks would need the removal rule to be supplemented by checks that do not wait for statistical evidence. We also note that skill text is derived from trajectories, so in any application where trajectories contain sensitive data, the library inherits that sensitivity and should be handled as derived data rather than as configuration.

\end{document}